\documentclass[10pt,twocolumn,letterpaper]{article}

\usepackage{iccv}
\usepackage{times}
\usepackage{epsfig}
\usepackage{graphicx}
\usepackage{amsmath}
\usepackage{amssymb}
\usepackage{color}
\usepackage{times}
\usepackage{overpic}
\usepackage{bm}
\usepackage{tabu}
\usepackage{bbding}
\usepackage{multicol}
\usepackage{multirow}
\usepackage[table]{xcolor}
\usepackage{comment}
\usepackage{graphicx}
\usepackage{times}
\usepackage{epsfig}
\usepackage{amsmath}
\usepackage{amssymb}
\usepackage{booktabs}
\usepackage{verbatim}
\usepackage{appendix}
\usepackage{subcaption}
\usepackage{adjustbox}
\usepackage{pifont}
\usepackage{enumitem}
\usepackage{makecell}
\usepackage{float,wrapfig}
\usepackage{wasysym}
\usepackage{nicefrac}

\PassOptionsToPackage{numbers, compress}{natbib}

\usepackage{color,xcolor}
\usepackage{epsfig}
\usepackage{graphicx}

\usepackage{array}
\usepackage{booktabs}
\usepackage{colortbl}
\usepackage{float,wrapfig}
\usepackage{hhline}
\usepackage{multirow}

\usepackage{amsmath,amsfonts,amsthm,amssymb}
\usepackage{bm}
\usepackage{nicefrac}
\usepackage{microtype}

\usepackage{changepage}
\usepackage{extramarks}
\usepackage{fancyhdr}
\usepackage{lastpage}
\usepackage{setspace}
\usepackage{soul}
\usepackage{xspace}
\usepackage{pifont}
\usepackage{cuted}
\usepackage{wrapfig}
\usepackage[table]{xcolor}
\usepackage{pifont}

\newcommand{\things}{\textit{things}}
\newcommand{\stuff}{\textit{stuff}}
\newcommand{\PQ}{PQ}
\newcommand{\PQda}{PQ\textsuperscript{$\dagger$}}
\newcommand{\RQ}{RQ}
\newcommand{\SQ}{SQ}
\newcommand{\PQth}{PQ\textsuperscript{Th}}
\newcommand{\RQth}{RQ\textsuperscript{Th}}
\newcommand{\SQth}{SQ\textsuperscript{Th}}
\newcommand{\PQst}{PQ\textsuperscript{St}}
\newcommand{\RQst}{RQ\textsuperscript{St}}
\newcommand{\SQst}{SQ\textsuperscript{St}}
\newcommand{\miou}{mIoU}

\usepackage[accsupp]{axessibility}

\usepackage[pagebackref=true,breaklinks,colorlinks]{hyperref}

\usepackage[capitalize]{cleveref}
\crefname{section}{Sec.}{Secs.}
\Crefname{section}{Section}{Sections}
\Crefname{table}{Table}{Tables}
\crefname{table}{Tab.}{Tabs.}

\definecolor{correct}{RGB}{173, 173, 173}
\definecolor{incorrect}{RGB}{192, 0, 0}

\iccvfinalcopy

\begin{document}

\title{UniSeg: A Unified Multi-Modal LiDAR Segmentation Network\\and the OpenPCSeg Codebase}

\author{Youquan Liu$^{1,2,}$\footnotemark[1]\quad Runnan Chen$^{1,3,}$\quad Xin Li$^{1,4}$\quad Lingdong Kong$^{1,5}$\quad Yuchen Yang$^{1,6}$\\Zhaoyang Xia$^{1,6}$\quad Yeqi Bai$^{1,}$\footnotemark[2]\quad Xinge Zhu$^{7}$\quad Yuexin Ma$^{8}$\quad Yikang Li$^{1,}$\footnotemark[2]\quad Yu Qiao$^{1}$\quad Yuenan Hou$^{1,}$\footnotemark[2]
\\[0.12ex]
\small{
$^{1}$Shanghai AI Laboratory \quad
$^{2}$Hochschule Bremerhaven \quad
$^{3}$The University of Hong Kong \quad
$^{4}$East China Normal University}\\
\small{
$^{5}$National University of Singapore \quad
$^{6}$Fudan University \quad
$^{7}$The Chinese University of Hong Kong \quad
$^{8}$Shanghai Tech University
}
\\
}
\maketitle
\def\algorithmname{UniSeg}

\renewcommand{\thefootnote}{\fnsymbol{footnote}} 
\footnotetext[1] {Work performed during an internship at Shanghai AI Laboratory.}
\footnotetext[2] {Corresponding authors.}

\begin{abstract}
\vspace{-7px}
Point-, voxel-, and range-views are three representative forms of point clouds. All of them have accurate 3D measurements but lack color and texture information. RGB images are a natural complement to these point cloud views and fully utilizing the comprehensive information of them benefits more robust perceptions. In this paper, we present a unified multi-modal LiDAR segmentation network, termed \algorithmname, which leverages the information of RGB images and three views of the point cloud, and accomplishes semantic segmentation and panoptic segmentation simultaneously. Specifically, we first design the \textbf{L}earnable cross-\textbf{M}odal \textbf{A}ssociation (LMA) module to automatically fuse voxel-view and range-view features with image features, which fully utilize the rich semantic information of images and are robust to calibration errors. Then, the enhanced voxel-view and range-view features are transformed to the point space, where three views of point cloud features are further fused adaptively by the \textbf{L}earnable cross-\textbf{V}iew \textbf{A}ssociation module (LVA). Notably, \algorithmname~achieves promising results in three public benchmarks, \textit{i.e.}, SemanticKITTI, nuScenes, and Waymo Open Dataset (WOD); it ranks \textbf{1}st on \textbf{two} challenges of two benchmarks, including the LiDAR semantic segmentation challenge of nuScenes and panoptic segmentation challenges of SemanticKITTI. Besides, we construct the \textrm{OpenPCSeg} codebase, which is the \textbf{largest} and \textbf{most comprehensive} outdoor LiDAR segmentation codebase. It contains most of the popular outdoor LiDAR segmentation algorithms and provides \textbf{reproducible} implementations. The OpenPCSeg codebase will be made publicly available at \url{https://github.com/PJLab-ADG/PCSeg}.

\end{abstract}


\begin{figure*}[t]
\centering
\includegraphics[width=1.0\linewidth]{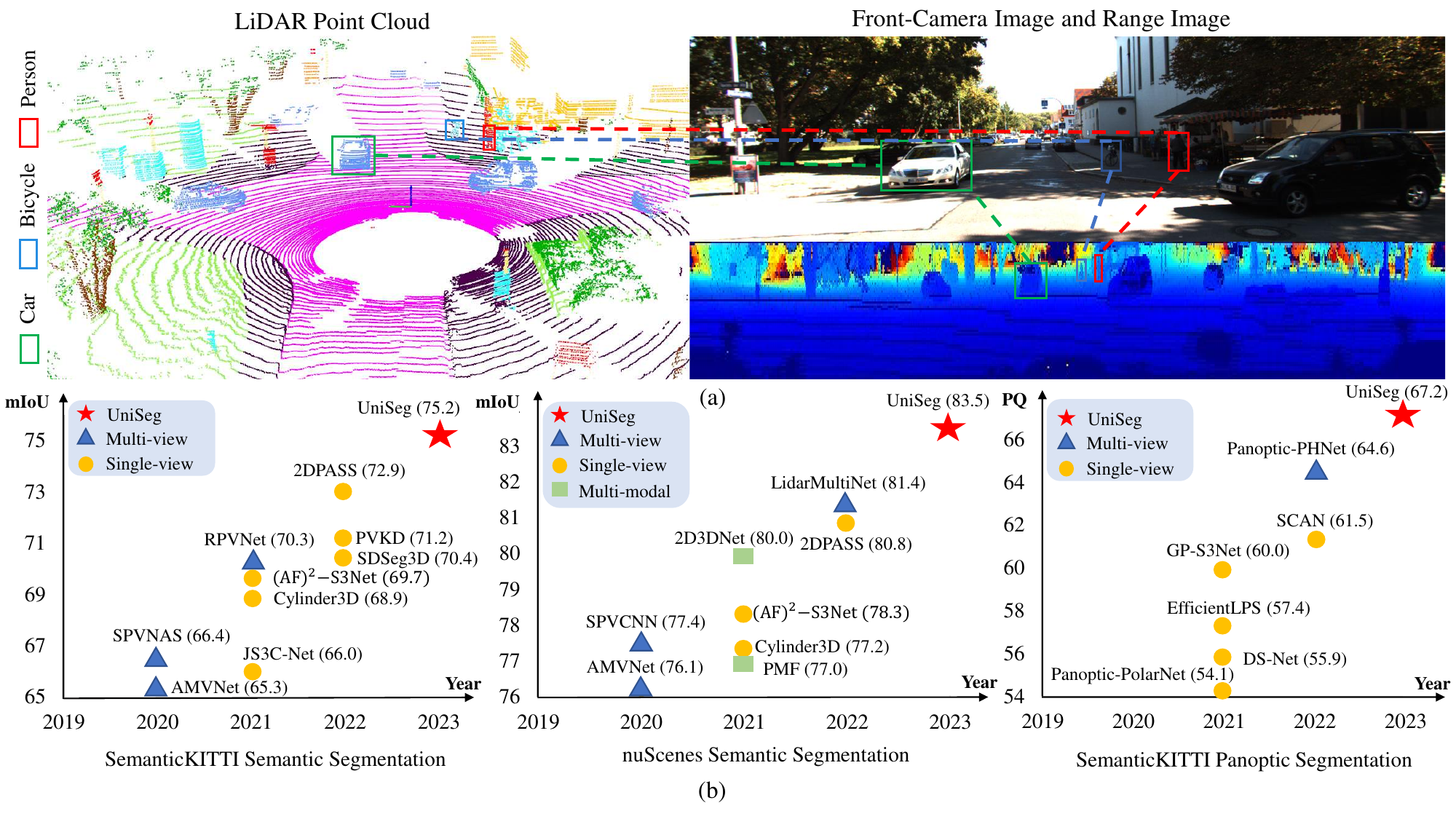}
\vskip -0.2cm
\caption{(a) Merits of different modalities. RGB images provide rich color, texture, and semantic information while point cloud embraces precise 3D positions of various objects. The pedestrian highlighted by the red rectangle is hard to find in the image but is visible in the point cloud. The combination of multi-modality and multi-views benefit a more robust and comprehensive perception. (b) Comparison of \algorithmname~with various competitive LiDAR segmentation algorithms on three challenges of SemanticKITTI and nuScenes benchmarks. The red pentagram, blue triangles, yellow circles, and green squares denote \algorithmname, multi-view methods, uni-modal methods, and multi-modal ones, respectively. The selected baselines include state-of-the-art algorithms such as 2DPASS~\cite{yan20222dpass}, RPVNet~\cite{rpvnet}, Panoptic-PHNet~\cite{li2022panoptic}, and LidarMultiNet~\cite{lidarmultinet}.}
\centering
\vspace{-1pt}
\label{fig:motivation}
\end{figure*}

\section{Introduction}
\label{sec:intro}
LiDAR-based semantic segmentation, whose objective is to assign a semantic label to each input point, acts as an essential component in autonomous driving, digital cities, and service robots \cite{3dseg_survey,3d_survey,huang2022multi,ma2023detzero}. With the advent of deep learning, an enormous amount of methods \cite{qi2017pointnet,zhu2021cylindrical,hmfi,logonet,pvkd2022,rpvnet,af2s3net,yan20222dpass,chen2023clip2Scene,chen2023cns,jiang2023deep,ji2022video} have been proposed and quickly dominate various benchmarks, such as SemanticKITTI~\cite{behley2019semantickitti} and nuScenes~\cite{caesar2020nuscenes,panoptic-nuscenes}.

Point cloud and RGB images are two frequently used modalities. As depicted in Fig.~\ref{fig:motivation} (a), different modalities have their own merits and drawbacks. Point cloud provides reliable and accurate depth information, and can be processed in different views, e.g., point-view, voxel-view, and range-view. Specifically, point-view representation maintains the complete point information but is inefficient in capturing the neighboring point features due to the unstructured point locations. Voxel-view methods rasterize the point cloud into voxel cells that retain regular structure but suffer from severe voxelization loss especially when the voxel size is large. Range-view representations are dense and compact, which can be efficiently processed by highly optimized 2D convolution. However, the spherical projection inevitably destroys the original 3D geometric information. As for the RGB image, it embraces rich color and texture information, but can not provide precise spatial information.

Apparently, the input data from multi-modality and multiple views of the point cloud are supplementary to each other. Therefore, fully utilizing the comprehensive information benefits a more robust perception. However, such a cross-modal and cross-view fusion paradigm is not fully explored in LiDAR segmentation~\cite{el2019rgb,krispel2020fuseseg,rpvnet,yan20222dpass}. Current multi-modal fusion methods are concentrated on the fusion of RGB and range images~\cite{el2019rgb,krispel2020fuseseg,kong2023rethinking}. Other representations such as voxel- and point-views of the LiDAR point cloud, which maintain original data structure and provide fine-grained spatial information, are ignored in prior methods. Besides, they typically fuse the image and point cloud in a hard association manner through calibration matrices, thus being vulnerable to calibration errors \cite{kong2023robo3D}.

In this paper, to address the aforementioned problems, we make the first attempt to dynamically fuse four different modalities of data (voxel-, range-, and point-views of the point cloud and RGB images) for more robust and accurate perception. More formally, we propose a Learnable cross-Modal Association (LMA) and a Learnable cross-View Association module (LVA) to effectively fuse the different modalities inputs. 
Specifically, we first fuse the image features with range- and voxel-view point features through the LMA in a soft association schema with the deformable cross-attention~\cite{zhu2020deformable} operation and alleviate calibration errors. Next, the image-enhanced range- and voxel-view features are transferred into the point-view feature, and all three views of point cloud features are fused adaptively by the LVA module.

Equipped with LMA and LVA, we design a unified network, dubbed \algorithmname, for various semantic scene understanding tasks, \textit{i.e.}, semantic, and panoptic segmentation. Extensive experimental results verify the generalizability of \algorithmname~across different tasks. As shown in Fig.~\ref{fig:motivation} (b), \algorithmname~ranks 1st in \textbf{two} open challenges. It achieves $75.2$ mIoU (semantic segmentation) and $67.2$ PQ (panoptic segmentation) in SemanticKITTI; and $83.5$ mIoU (semantic segmentation) and $78.4$ PQ (panoptic segmentation) in nuScenes. The appealing performance strongly demonstrates the efficacy of our multi-modal fusion framework.

Besides, considering that many popular outdoor LiDAR segmentation methods~\cite{af2s3net,rpvnet,pvkd2022,li2022panoptic} either do not provide official implementations or the performance is difficult to reproduce, we construct the OpenPCSeg codebase which aims to provide reproducible and uniform implementations. We have benchmarked $\mathbf{14}$ competitive LiDAR segmentation algorithms and the reproduced performance of these algorithms all surpasses the reported value.

The contributions of our work are summarized as follows.
\begin{itemize}

\item {We propose a unified multi-modal fusion network for LiDAR segmentation, leveraging the information of RGB images and three views of the point cloud for more accurate and robust perception.}
\item {Our approach ranks $1$st on two challenges of SemanticKITTI and nuScenes, strongly demonstrating the efficacy of the proposed multi-modal network.}
\item {The largest and most comprehensive outdoor LiDAR segmentation codebase dubbed OpenPCSeg will be released to facilitate related research.}
\end{itemize}

\section{Related Work}
\label{sec:related}
\subsection{LiDAR-Based Semantic Scene Understanding}
Semantic segmentation \cite{rpvnet,af2s3net,pvkd2022,zhu2021cylindrical,zhu2021cylindrical-tpami,tang2020searching,yan20222dpass,lasermix,kong2023conDA,chen2022zero,chen2022towards,chen2020unsupervised,liu2023segment,lu2023see,xu2023human,kong2023rethinking} and panoptic segmentation \cite{dsnet,li2022panoptic}  are two basic tasks for LiDAR-based semantic scene understanding. LiDAR semantic segmentation aims to assign a class label to each point in the input point cloud sequence. LiDAR panoptic segmentation performs semantic segmentation and instance segmentation on the stuff class and thing class, respectively. The majority of the LiDAR segmentation approaches take the point cloud as the sole input signal. For instance, Cylinder3D \cite{zhu2021cylindrical,zhu2021cylindrical-tpami,pvkd2022} divides the point cloud with cylindrical partition and feeds these cylinder features into the UNet-based segmentation backbone. SPVCNN~\cite{tang2020searching} introduces the point branch to complement the original voxel branch and performs pointwise segmentation based on the fused point-voxel features. LidarMultiNet~\cite{lidarmultinet} unifies LiDAR semantic segmentation, panoptic segmentation, and 3D object detection in one network and achieves impressive perception performance. The preceding methods ignore the rich information contained in RGB images, thus yielding sub-optimal performance. On the contrary, our \algorithmname~takes all modalities and all views of the point cloud into account and can benefit from the merits of all input signals. 

\subsection{Multi-Modal Sensor Fusion}
Since the uni-modal signal has its own shortcomings, multi-modal fusion is gaining increasing attention in recent years~\cite{zhuang2021perception,el2019rgb,krispel2020fuseseg}. Zhuang \etal ~\cite{zhuang2021perception} projects the point cloud into the perspective view and fuses the multi-modal features through the residual-based fusion module. El Madawi \etal ~\cite{el2019rgb} performs early fusion and middle fusion of the range images and re-projected RGB images. Krispel \etal ~\cite{krispel2020fuseseg} incorporates the image features into the range-image-based backbone via the calibration matrices. The above-mentioned approaches merely perform one-to-one multi-modal fusion and cannot fully utilize the rich semantic information of RGB images. And these methods yield inferior performance when the calibration matrices are inaccurate. By contrast, our method can achieve more adaptive multi-modal feature fusion and relieve point-pixel misalignment using the proposed learnable cross-modal association module.

\section{The OpenPCSeg Codebase}
In the outdoor LiDAR segmentation field, many popular semantic segmentation algorithms~\cite{af2s3net,rpvnet,li2022panoptic,pvkd2022} either do not release their official implementations or the released codes are difficult to reproduce the reported performance. Currently, only a few open-sourced projects have provided the implementations of LiDAR segmentation models such as the well-known mmdetection3d project~\cite{mmdet3d2020}. However, it only includes some classical indoor LiDAR segmentation algorithms. A brief comparison between mmdetection3d and our OpenPCSeg is presented in Table~\ref{codebase_table}. To facilitate the research in the outdoor LiDAR segmentation area, we construct the largest and most comprehensive OpenPCSeg codebase that contains the reproducible implementations of these competitive LiDAR segmentation models. OpenPCSeg is built upon the noted OpenPCDet~\cite{openpcdet} project. 
Considering the fact that many implementation details are missing in the original paper, constructing such a codebase is non-trivial. It takes us around one year to build the codebase through an enormous number of experiments to determine the optimal selection of hyperparameters, data augmentations, optimizers, learning rate schedules, data pre-processing, and post-processing strategies, \etc. Till now, we have successfully reproduced more than ten competitive outdoor LiDAR segmentation algorithms, such as SalsaNext~\cite{cortinhal2020salsanext}, Cylinder3D~\cite{zhu2021cylindrical}, RPVNet~\cite{rpvnet} and SPVCNN~\cite{liu2019point}. The reproduced performance of these algorithms all surpasses the reported value in their original publications. The chosen datasets include SemanticKITTI~\cite{behley2019semantickitti} and nuScenes~\cite{caesar2020nuscenes,panoptic-nuscenes}. The selected tasks contain LiDAR semantic segmentation and panoptic segmentation. We provide a full suite of training and inference protocols for these algorithms to ensure reproducibility. The complete performance comparison and additional information on the OpenPCSeg codebase are in the Appendix.

\begin{table}[!t]
\caption{Comparisons between existing codebase.}
\vskip -0.2cm
\label{codebase_table}
\centering
\scalebox{0.87}{
\begin{tabular}{c|c|c|c}
\toprule
Codebase & Task & Task Difficulty & \#Method \\
\midrule
MMDetection3D & Indoor Seg & Relatively Easy & $3$  \\
\rowcolor{cyan!9}\textbf{OpenPCSeg} & Outdoor Seg & Difficult & $\mathbf{14}$  \\
\bottomrule
\end{tabular}
}
\vspace{-2ex}
\end{table}
\begin{figure*}[!ht]
 \centering
 \includegraphics[width=1.0\linewidth]{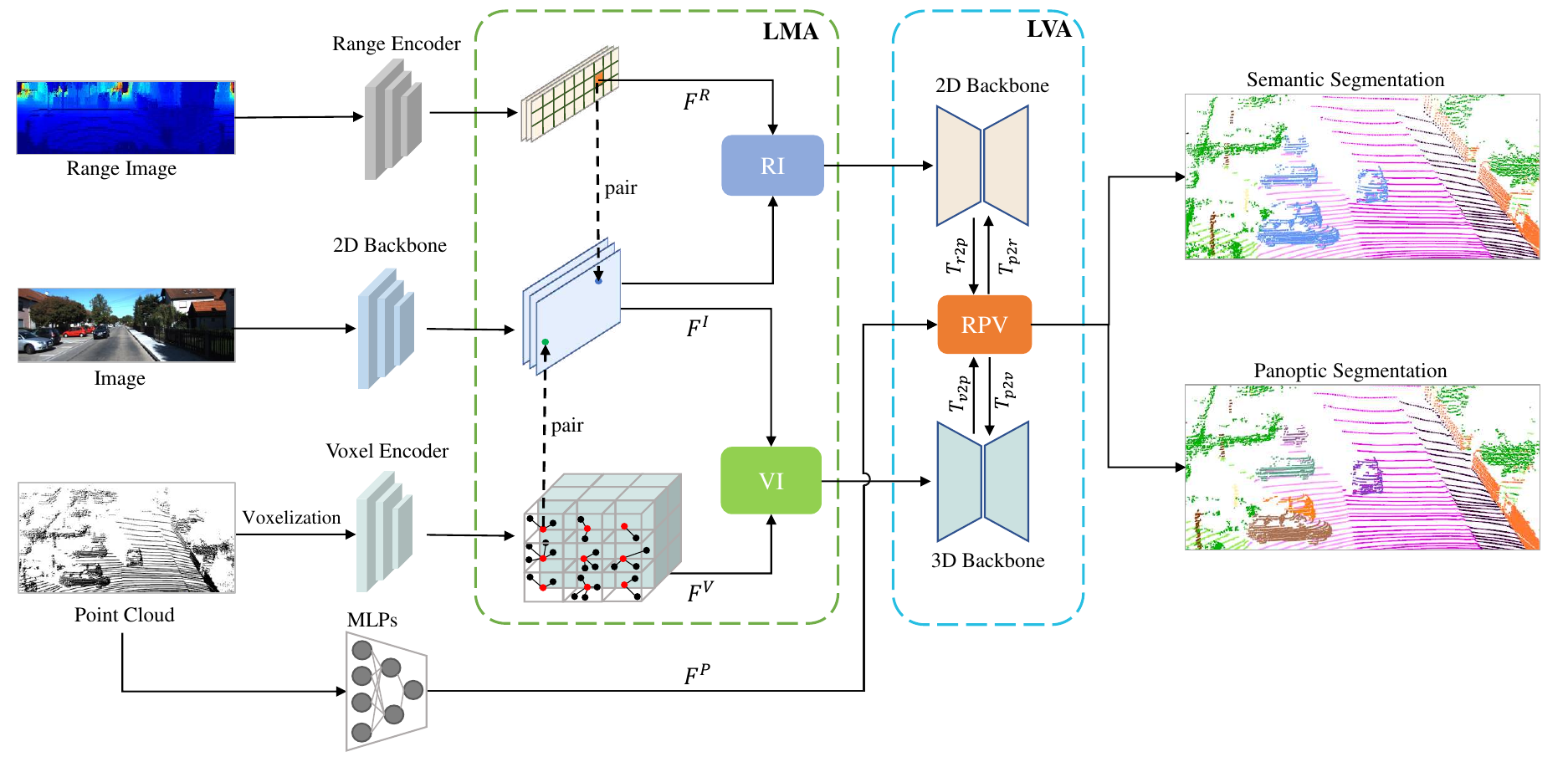}
 \vskip -0.2cm
 \caption{\textbf{Framework overview}. \algorithmname~takes four signals as input, \ie, point cloud (voxel-, range- and point-views) and RGB images. Given the input point cloud, the range-, voxel-, and point-view features are produced by a 2D range encoder, a 3D voxel encoder, and MLPs, respectively. For the voxel features and range image features, we fuse them with the RGB image features (VI and RI) via the proposed learnable cross-modal association (LMA) module. Then, for the range image features and voxel features, we project them to the point space via the range-image-to-point transformation $T_{r2p}$ and voxel-to-point transformation $T_{v2p}$. Features of these three views of the point cloud are fused (RPV) by the learnable cross-view association (LVA) module and we perform fusion at different layers to leverage both low-level and high-level information.}
 \centering
 \vskip -0.2cm
 \label{fig:framework}
\end{figure*}

\section{Methodology}
\label{sec:methodology}
\subsection{Framework Overview}
\algorithmname~takes point cloud (voxel-, range- and point-views) and RGB images as input and performs semantic segmentation and panoptic segmentation in a single network. Specifically, the input point cloud is $\mathbf{X} \in \mathbb{R}^{N \times 3}$ and the input image is $I \in \mathbb{R}^{H \times W \times 3}$. $N$ is the number of points, $H$ and $W$ are the height and width of the image, respectively. We obtain the range image representation by performing the spherical projection on the point cloud. The range image is fed to a range-view-based backbone to extract range image features $\mathbf{F}^{R} \in \mathbb{R}^{H_{R} \times W_{R} \times C_{R}}$. $H_{R}$, $W_{R}$, and $C_{R}$ are the height, width, and number of channels of the range image feature, respectively. Then, we extract the point features $\mathbf{F}^{P} \in \mathbb{R}^{N \times C_{p}}$ via a series of Multi-Layer Perceptrons (MLPs), where $C_{p}$ is the number of channels of the point features. The voxel features $\mathbf{F}^{V} \in \mathbb{R}^{N_{v} \times C_{p}}$ are produced by the voxelization process that performs max pooling on the point features in one voxel. $N_{v}$ is the number of non-empty voxels. The input image is fed to a ResNet-based architecture to extract the image features $\mathbf{F}^{I} \in \mathbb{R}^{H_{I} \times W_{I} \times C_{I}}$. $H_{I}$, $W_{I}$, and $C_{I}$ are the height, width, and number of channels of the image feature, respectively.

Our method consists of two modules, \ie, Learnable cross-Modal Association (LMA) and Learnable cross-View Association (LVA). The LMA module copes with the voxel-image fusion and range-image fusion, and the LVA module concentrates on range-point-voxel fusion. In what follows, we present LMA and LVA in detail.

\subsection{Learnable Cross-Modal Association}

\noindent\textbf{Point-Image Calibration}.
We build the correspondence between the points and RGB image pixels via camera calibration matrices. Specifically, for each point coordinate $(x_i, y_i, z_i)$, the corresponding pixel $(u_i, v_i)$ is found by the following:
\begin{equation}
\label{eqn:voxel_pixel_pair}
\begin{split}
[u_i, v_i, 1]^\text{T} = \frac{1}{z_i}  \cdot S\cdot  T \cdot [x_i, y_i, z_i, 1]^\text{T},
\end{split}
\end{equation}
where $T \in \mathbb{R}^{4\times4}$ is the camera extrinsic matrix that consists of a rotation matrix and a translation matrix, and $S \in \mathbb{R}^{3\times4}$ is the camera intrinsic matrix. Here, we denote this pixel $(u_i, v_i)$ as calibrated pixel $p_i$ and the corresponding image feature as calibrated image feature $F_i^I$.

\begin{figure}
\centering
\includegraphics[width=1.0\linewidth]{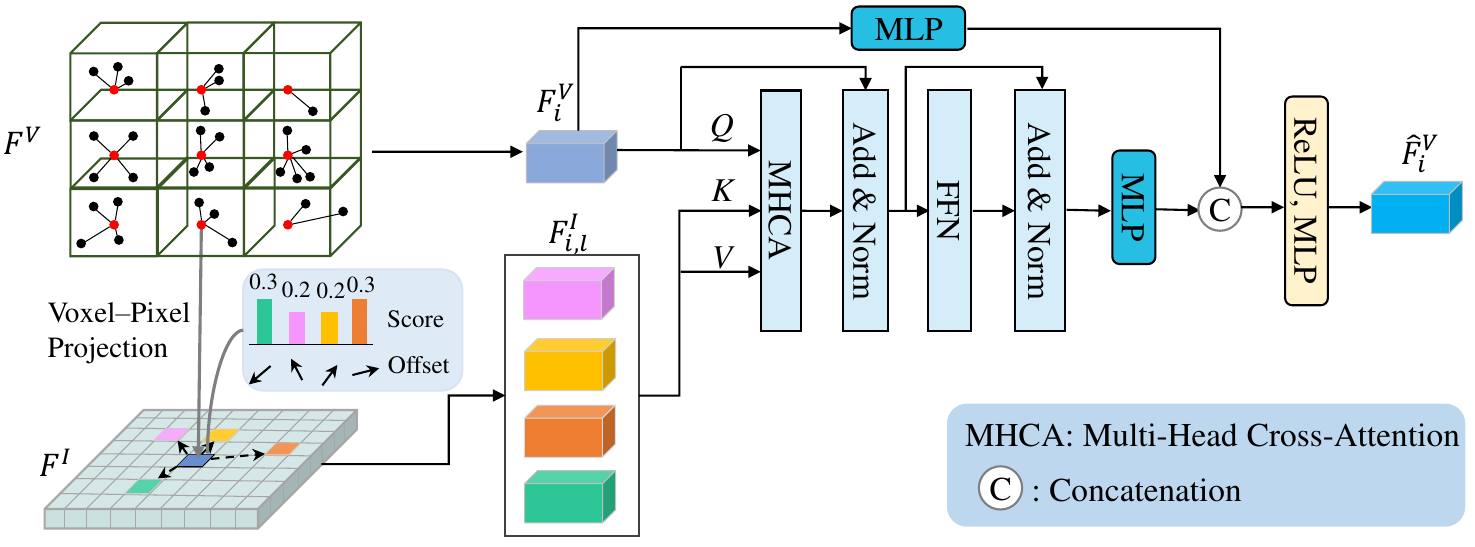}
\vskip -0.1cm
\caption{\textbf{Voxel-Image Fusion}. For each voxel feature $F^V_i$, We first calculate the calibrated image feature $F^I_i$ based on the voxel center and calibration
matrices. Then, we leverage learned offsets to sample $L$ image features. The voxel feature is treated as \textit{Query}, and the sampled image features are denoted as \textit{Key} and \textit{Value}. The voxel and sampled image features are fed to the multi-head cross-attention module to obtain image-enhanced voxel features. These features are concatenated with the original features to produce the final fused features.}
\centering
\vspace{-3pt}
\label{fig:vi}
\end{figure}

\noindent\textbf{Voxel-Image Fusion}.
Previous multi-modal fusion approaches~\cite{el2019rgb,krispel2020fuseseg} heavily rely on imperfect camera calibration matrices, which are vulnerable to calibration errors. Inspired by deformable detr~\cite{deformabledetr}, we adaptively fuse the voxel features with image features to alleviate the calibration errors. As shown in Fig.~\ref{fig:vi}, the voxel coordinate is the voxel centre, and the calibrated image pixel is calculated by Equation \ref{eqn:voxel_pixel_pair}. Next, we estimate the image pixel offsets from the calibrated image pixel, and then we fuse the selected image feature with the corresponding voxel feature as follows:
\begin{equation}
\begin{array}{cc}
F^I_{i, l} = F^I({{\bf{p}}_i} + \Delta {{\bf{p}}_{i, l}}),\\
{{\hat F}^V_i} = \sum\limits_{m = 1}^M {{W_m}} \left[ \sum\limits_{l = 1}^L {{A_{i, l, m}}\cdot(W_m^{'}{ F^I_{i, l}})} \right],
\end{array}
\label{eqn:vi-fuse}
\end{equation}
where $F^I$ is the image feature, $F_{i, l}^I$ are the sampled image features and ${\hat F}_i^V$ is the image-enhanced voxel feature. $W_m$ and $W_m^{'}$ are the learnable weights, $m$ indexes the attention head, $M$ is the number of self-attention heads and $L$ is the total number of sampled image features. $\Delta {{\bf{p}}_{i, l}}$ and $A_{i, l, m}$ denote the sampling offset and attention weight of the $l$-th sampled image feature in the $m$-th attention head, respectively. Both are obtained by performing the linear projection on the voxel feature $F^V_i$. We concatenate the image-enhanced voxel feature ${\hat F}^V_i$ with the original voxel feature to obtain the final fused voxel feature $\hat F^V_i \in \mathbb{R}^{N \times2C_f}$, where $C_f$ is the number of channels of the voxel feature. Therefore, the voxel feature will automatically find the most relevant image features to fuse. Note that those voxel features that do not have the corresponding image features will be appended with zero vectors.

\noindent \textbf{Range-Image Fusion}. As to the range-image fusion, we follow the same process with voxel-image fusion (Equation \ref{eqn:vi-fuse}), thus producing the final image-enhanced range-view features $\hat{F}^{R} \in \mathbb{R}^{H_{R} \times W_{R} \times C_{f}}$.

\subsection{Learnable Cross-View Association}

\begin{figure}[!t]
\centering
\includegraphics[width=1.0\linewidth]{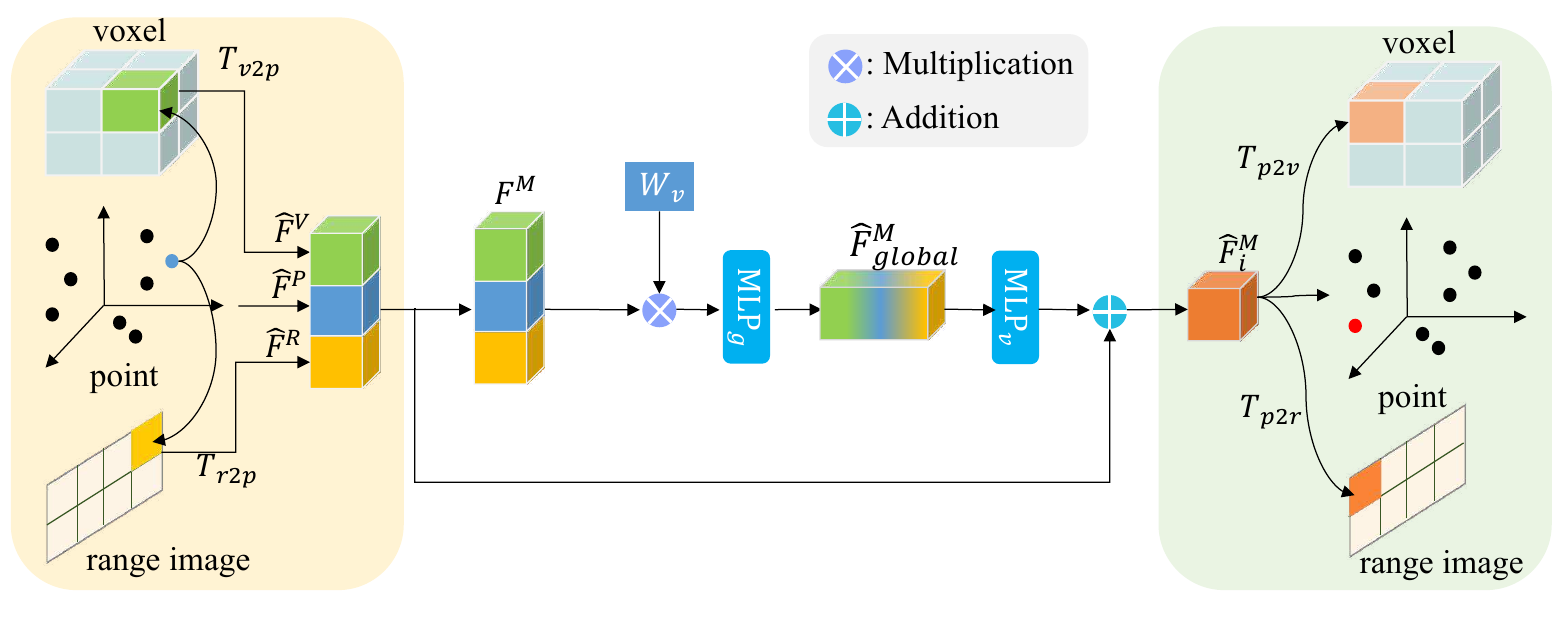}
\vskip -0.1cm
\caption{\textbf{Learnable cross-View Association (LVA)}. Voxel and range image features are first mapped to the point space where interpolations are employed to address the quantity mismatch problem through $T_{v2p}$ and $T_{r2p}$ transformations. 
Given voxel-, point- and range-view features, the LVA extracts its global representation and view-wised adapted features. Via a residual connection, the cross-view fused feature is obtained and projected back to the original voxel and range image space through $T_{p2v}$ and $T_{p2r}$ transformations.}
\centering
\vspace{-3pt}
\label{fig:rpv}
\end{figure}

After the learnable cross-modal association module, we obtain the image-enhanced voxel- and range-view features. For the range-, point-, voxel-view features fusion, we first apply the range-to-point transformation $T_{r2p}$ and voxel-to-point transformation $T_{v2p}$ on the range-, voxel-view features to transfer them into the point-view respectively. And we propose a learnable cross-view association module to dynamically integrate these three modalities' features, as shown in Fig.~\ref{fig:rpv}.

Specifically, in the $T_{r2p}$ and $T_{v2p}$ transformations, since the number of voxel features and range image features is smaller than the number of points, directly appending all-zero vectors to voxel features and range image features yields sub-optimal performance. To address the aforementioned quantity mismatch problem, we resort to trilinear interpolation and bilinear interpolation to generate interpolated voxel features and pseudo range image features, respectively.

After these transformations, we obtain the point-wise voxel features $\hat{F}^V \in \mathbb{R}^{m \times C_{f}}$, point-wise range image features $\hat{F}^R  \in \mathbb{R}^{m \times C_{f}}$ and point features $\hat{F}^P  \in \mathbb{R}^{m \times C_{f}}$. And we concatenate them to produce the multi-view feature $F^M \in \mathbb{R}^{m \times 3C_{f}}$. Then $F^M$ is weighted by the learnable parameters $W_v$ and obtains the compact global point feature via the first two layers of LVA, \ie, $\mathrm{MLP}_g$, as follows:
\begin{equation}
\begin{array}{c}
\hat{F}_{global}^M = \mathrm{ReLU}(\mathrm{MLP}_{g}(W_v(\mathrm{concat}(\hat{F}^V, \hat{F}^R, \hat{F}^P)))),
\end{array}
\label{eqn:global}
\end{equation}
where $\hat{F}_{global}^M \in \mathbb{R}^{m \times C_{f}}$.Through this cross-view aggregation, multi-view features fuse into a summative representation. After that, the view-wise adapted feature is generated from the globally enhanced features $\hat{F}_{global}^M$ and adds its original features of different views which are obtained by a residual connection as follows:
\begin{equation}
\begin{array}{c}
{\hat F_i}^M = \hat{F}_{i} + \mathrm{ReLU}(\mathrm{MLP}_{v}(\hat{F}_{global}^M)),
\end{array}
\label{eqn:view_wise}
\end{equation}
where $\hat{F}_{i} \in \mathbb{R}^{m \times C_{f}} $ denotes the original feature in point space for view $i$. On the one hand, $ {\hat F_i}^M$ provides global adapted features into $\hat{F}_{i}$ for a better representation of three different views. On the other hand, the residual style combines the benefits of multi-view knowledge with those of its advantages, which further encourages cross-view interaction. The final cross-view feature ${\hat F_i}^M$ is projected back to the original voxel and range image space by the $T_{p2v}$ and $T_{p2r}$ transformations respectively.

\subsection{Task-Specific Heads}

The fused features obtained by the LMA and LVA modules will be fed to the classifier to produce the semantic segmentation predictions. The semantic predictions are passed to the panoptic head to estimate instance centre positions and offsets of the thing class, producing the panoptic segmentation results. Detailed panoptic segmentation implementation is described in the supplementary material.

\begin{table*}[ht]
\caption{Quantitative results of \algorithmname~and SoTA LiDAR semantic segmentation methods on the SemanticKITTI \textit{test} set.}
\vskip -0.3cm
\label{tab:sem_ss}
\centering
\begin{adjustbox}{width=\textwidth}
\begin{tabular}{r|c|c|c|c|c|c|c|c|c|c|c|c|c|c|c|c|c|c|c|c}
\toprule
\textbf{Method} & \textbf{mIoU} & \rotatebox{0}{car} & \rotatebox{0}{bicy} & \rotatebox{0}{moto} & \rotatebox{0}{truc} & \rotatebox{0}{o.veh} & \rotatebox{0}{ped} & \rotatebox{0}{b.list} & \rotatebox{0}{m.list} & \rotatebox{0}{road} & \rotatebox{0}{park} & \rotatebox{0}{walk} & \rotatebox{0}{o.gro} & \rotatebox{0}{build} & \rotatebox{0}{fenc} & \rotatebox{0}{veg} & \rotatebox{0}{trun} & \rotatebox{0}{terr} & \rotatebox{0}{pole} & \rotatebox{0}{sign}
\\
\midrule
AMVNet~\cite{liong2020amvnet} & $65.3$ & $96.2$ & $59.9$ & $54.2$ & $48.8$ & $45.7$ & $71.0$ & $65.7$ & $11.0$ & $90.1$ & $71.0$ & $75.8$ & $32.4$ & $92.4$ & $69.1$ & $85.6$ & $71.7$ & $69.6$ & $62.7$ & $67.2$
\\
JS3C-Net~\cite{yan2021sparse} & $66.0$ & $95.8$ & $59.3$ & $52.9$ & $54.3$ & $46.0$ & $69.5$ & $65.4$ & $39.9$ & $88.9$ & $61.9$ & $72.1$ & $31.9$ & $92.5$ & $70.8$ & $84.5$ & $69.8$ & $67.9$ & $60.7$ & $68.7$ 
\\
SPVNAS~\cite{tang2020searching} & $66.4$ & $97.3$ & $51.5$ & $50.8$ & $59.8$ & $58.8$ & $65.7$ & $65.2$ & $43.7$ & $90.2$ & $67.6$ & $75.2$ & $16.9$ & $91.3$ & $65.9$ & $86.1$ & $73.4$ & $71.0$ & $64.2$ & $66.9$
\\
Cylinder3D~\cite{zhu2021cylindrical} & $68.9$ & $97.1$ & $67.6$ & $63.8$ & $50.8$ & $58.5$ & $73.7$ & $69.2$ & $48.0$ & $92.2$ & $65.0$ & $77.0$ & $32.3$ & $90.7$ & $66.5$ & $85.6$ & $72.5$ & $69.8$ & $62.4$ & $66.2$
\\
AF2S3Net~\cite{af2s3net} & $69.7$ & $94.5$ & $65.4$ & $\mathbf{86.8}$ & $39.2$ & $41.1$ & $\mathbf{80.7}$ & $80.4$ & $\mathbf{74.3}$ & $91.3$ & $68.8$ & $72.5$ & $\mathbf{53.5}$ & $87.9$ & $63.2$ & $70.2$ & $68.5$ & $53.7$ & $61.5$ & $71.0$
\\
RPVNet~\cite{rpvnet} & $70.3$ & $97.6$ & $68.4$ & $68.7$ & $44.2$ & $61.1$ & $75.9$ & $74.4$ & $73.4$ & $\mathbf{93.4}$ & $70.3$ & $\mathbf{80.7}$ & $33.3$ & $\mathbf{93.5}$ & $72.1$ & $86.5$ & $75.1$ & $71.7$ & $64.8$ & $61.4$
\\
SDSeg3D~\cite{li2022self} & $70.4 $ & $97.4$ & $58.7$ & $54.2$ & $54.9$ & $65.2$ & $70.2$ & $74.4$ & $52.2$ & $90.9$ & $69.4$ & $76.7$ & $ 41.9$ & $93.2$ & $71.1$ & $86.1$ & $74.3$ & $71.1$ & $65.4$ & $70.6$
\\
GASN~\cite{ye2022efficient} & $70.7$ & $96.9$ & $65.8$ & $ 58.0$ & $59.3$ & $61.0$ & $80.4$ & $\mathbf{82.7}$ & $46.3$ & $89.8$ & $66.2$ & $74.6$ & $30.1$ & $92.3$ & $69.6$ & $87.3$ & $73.0$ & $72.5$ & $66.1$ & $\mathbf{71.6}$
\\
PVKD~\cite{pvkd2022} & $71.2$ & $97.0$ & $67.9$ & $69.3$ & $53.5$ & $60.2$ & $75.1$ & $73.5$ & $50.5$ & $91.8$ & $70.9$ & $77.5$ & $41.0$ & $92.4$ & $69.4$ & $86.5$ & $73.8$ & $71.9$ & $64.9$ & $65.8$
\\
2DPASS~\cite{yan20222dpass} & $72.9$ & $97.0$ & $63.6$ & $63.4$ & $61.1$ & $61.5$ & $77.9$ & $81.3$ & $74.1$ & $89.7$ & $67.4$ & $74.7$ & $40.0$ & $\mathbf{93.5}$ & $\mathbf{72.9}$ & $86.2$ & $73.9$ & $71.0$ & $65.0$ & $70.4$
\\
RangeFormer~\cite{kong2023rethinking} & $73.3$ & $96.7$ & $69.4$ & $73.7$ & $59.9$ & $66.2$ & $78.1$ & $75.9$ & $58.1$ & $92.4$ & $73.0$ & $78.8$ & $42.4$ & $92.3$ & $70.1$ & $86.6$ & $73.3$ & $72.8$ & $66.4$ & $66.6$
\\
\midrule
\rowcolor{cyan!9}\textbf{\algorithmname~(Ours)} & $\mathbf{75.2}$ & $\mathbf{97.9}$ & $\mathbf{71.9}$ & $75.2$ & $\mathbf{63.6}$ & $\mathbf{74.1}$ & $78.9$ & $74.8$ & $60.6$ & $92.6$ & $\mathbf{74.0}$ & $79.5$ & $46.1$ & $93.4$ & $72.7$ & $\mathbf{87.5}$ & $\mathbf{76.3}$ & $\mathbf{73.1}$ & $\mathbf{68.3}$ & $68.5$
\\\bottomrule
\end{tabular}
\end{adjustbox}
\end{table*}

\begin{table*}[!t]
\caption{Quantitative results of \algorithmname~and SoTA LiDAR semantic segmentation methods on the nuScenes \textit{test} set.}
\vskip -0.3cm
\label{tab:nusc}
\centering
\begin{adjustbox}{width=\textwidth}
\begin{tabular}{r|c|c|c|c|c|c|c|c|c|c|c|c|c|c|c|c|c}
\toprule
\textbf{Method} & \textbf{mIoU} & \rotatebox{0}{barr} &  \rotatebox{0}{bicy} & \rotatebox{0}{bus} & \rotatebox{0}{car} & \rotatebox{0}{const} & \rotatebox{0}{motor} & \rotatebox{0}{ped} & \rotatebox{0}{cone} & \rotatebox{0}{trail} & \rotatebox{0}{truck} & \rotatebox{0}{driv} & \rotatebox{0}{other} &
\rotatebox{0}{walk} & \rotatebox{0}{terr} & \rotatebox{0}{made} & \rotatebox{0}{veg}
\\\midrule
PMF~\cite{zhuang2021perception}
&77.0 & 82.0& 40.0& 81.0& 88.0& 64.0 &79.0& 80.0 & 76.0& 81.0& 67.0& 97.0& 68.0& 78.0& 74.0& 90.0& 88.0\\
Cylinder3D~\cite{zhu2021cylindrical} & 77.2 & 82.8 & 29.8 & 84.3 & 89.4 & 63.0 & 79.3 & 77.2 & 73.4 & 84.6 & 69.1 & 97.7 & 70.2 & 80.3 & 75.5 & 90.4 & 87.6  \\

AMVNet~\cite{liong2020amvnet} &77.3 &  80.6 & 32.0 & 81.7 & 88.9 & 67.1 & 84.3 & 76.1 & 73.5 & 84.9 & 67.3 & 97.5 & 67.4 & 79.4 & 75.5 & 91.5 & 88.7\\

SPVCNN~\cite{tang2020searching} &77.4& 80.0& 30.0 & 91.9 & 90.8 & 64.7 & 79.0 & 75.6 & 70.9 & 81.0 & 74.6 & 97.4 & 69.2 & 80.0 & 76.1 & 89.3 & 87.1 \\
AF2S3Net~\cite{af2s3net} &78.3 & 78.9 & 52.2 & 89.9 & 84.2 & 77.4 & 74.3 & 77.3 & 72.0 & 83.9 & 73.8 & 97.1 & 66.5 & 77.5 & 74.0 & 87.7 & 86.8 \\
2D3DNet~\cite{genova2021learning} & 80.0 & 83.0 & 59.4 & 88.0 &85.1 & 63.7 & 84.4 & 82.0 & 76.0 & 84.8 & 71.9 & 96.9 & 67.4 & 79.8 & 76.0 & \bf{92.1} & 89.2 \\
GASN~\cite{ye2022efficient} &80.4 & 85.5 & 43.2 & 90.5 & \bf{92.1} & 64.7 & 86.0 & 83.0 & 73.3 & 83.9 & 75.8 & 97.0 & 71.0 & \bf{81.0} & \bf{77.7} & 91.6 & \bf{90.2}\\
2DPASS~\cite{yan20222dpass} & 80.8 & 81.7 & 55.3 & 92.0 & 91.8 & 73.3 & 86.5 & 78.5 & 72.5 & 84.7 & 75.5 & 97.6 & 69.1 & 79.9 & 75.5 & 90.2 & 88.0\\
LidarMultiNet~\cite{lidarmultinet} & 81.4 & 80.4 & 48.4 & \bf{94.3} & 90.0 & 71.5 & 87.2 & \bf{85.2} & \bf{80.4} & \bf{86.9} & 74.8 & \bf{97.8} & 67.3 & 80.7 & 76.5 & \bf{92.1} & 89.6 \\
\midrule
\rowcolor{cyan!9}\textbf{\algorithmname~(Ours)} & \bf{83.5} & \bf{85.9} & \bf{71.2} & 92.1 & 91.6 & \bf{80.5} & \bf{88.0} & 80.9 & 76.0 & 86.3 & \bf{76.7} & 97.7 & \bf{71.8} & 80.7 & 76.7 &  91.3 & 88.8
\\\bottomrule
\end{tabular}
\end{adjustbox}
\end{table*}

\subsection{Overall Objective}

The overall loss function is comprised of four terms, \ie, the cross-entropy loss, the Lovasz-softmax loss~\cite{berman2018the}, the heatmap regression via MSE loss, and the offset map regression by L1 loss, \textit{i.e.},
\begin{equation}
\label{eqn:loss_func}
\begin{split}
\mathcal{L} = & \mathcal{L}_{\mathrm{wce}} + \alpha \mathcal{L}_{\mathrm{lovasz}} + \beta \mathcal{L}_{\mathrm{heatmap}} + \gamma \mathcal{L}_{\mathrm{offset}},
\end{split}
\end{equation}

\noindent where $\alpha$, $\beta$, and $\gamma$ are the loss coefficients to balance the effect of each loss term.

\section{Experiments}
\label{sec:seg}
\noindent \textbf{Datasets}. Following the practice of popular LiDAR segmentation models \cite{zhu2021cylindrical,dsnet,pvkd2022}, we conduct experiments on three popular benchmarks, \ie, nuScenes \cite{caesar2020nuscenes,panoptic-nuscenes}, SemanticKITTI \cite{behley2019semantickitti}, and Waymo Open \cite{waymo}. For nuScenes, it consists of $1000$ driving scenes where 850 scenes are selected for training and validation, and the remaining $150$ scenes are taken as the testing split. $16$ classes are utilized for LiDAR semantic segmentation after merging similar classes and eliminating infrequent classes. As to SemanticKITTI, it has $22$ point cloud sequences. Sequences $00$ to $10$, $08$, and $11$ to $21$ are used for training, validation, and testing, respectively. $19$ classes are chosen for training and evaluation after merging classes with distinct moving statuses and discarding classes with very few points. The Waymo Open Dataset (WOD) has $798$, $202$, and $150$ sequences for training, validation, and testing, respectively. The duration of each sequence is $20$ seconds and the frame rate is $10$ Hz. However, for the 3D semantic segmentation task, not all frames are provided with 3D segmentation annotations. Specifically, only the last frame of a fixed number of frames is annotated. The number of annotated frames for training and validation is $23691$, and $5976$, respectively. The total number of classes is $23$, including one ignored and $22$ valid semantic categories. Note that both the first return and second return of the point cloud need to be segmented.

\noindent \textbf{Evaluation Metrics}. Following the practice of~\cite{pvkd2022,zhu2021cylindrical}, we adopt the Intersection-over-Union (IoU) of each class and mIoU of all classes as the evaluation metric. The IoU of class $i$ is calculated via $IoU_{i} = \frac{TP_{i}}{TP_{i}+FP_{i}+FN_{i}}$, where $TP_{i}$, $FP_{i}$ and $FN_{i}$ denote the true positive, false positive and false negative of class $i$, respectively. For panoptic segmentation, we adopt the Panoptic Quality (PQ) as the main metric.

\noindent \textbf{Implementation Details}. For the point cloud branch, we first construct the point-voxel backbone based on the Minkowski-UNet34~\cite{choy20194d}. Then, we add the range-image branch, \ie, SalsaNext~\cite{cortinhal2020salsanext}, to the point-voxel network and perform point-voxel-range fusion at four levels. The number of training epochs is set as 36 and the initial learning rate is set as 0.12. We use SGD as the optimizer. We use 1 epoch to warm up the network and adopt the cosine learning rate schedule for the remaining epochs. The momentum is set at 0.9 and weight decay is set at 0.0001. The voxel size is set as 0.05 for SemanticKITTI and WOD, and 0.1 for nuScenes. The gradient norm clip is set to 10 to stabilize the training process. $\alpha$, $\beta$ and $\gamma$ are set as 1, 100, and 10, respectively. As to data augmentation of the point cloud branch, we employ random flip, random scaling, random translation as well as LaserMix~\cite{lasermix} and PolarMix~\cite{xiao2022polarmix} to increase the diversity of training samples. For the RGB image branch, we use ImageNet-pretrained ResNet-34 as the feature extractor. The parameters in the image branch are trainable. More details are put in the supplementary.

\begin{table}[!t]
  \caption{Quantitative results of \algorithmname~and SoTA LiDAR panoptic segmentation methods on SemanticKITTI \textit{test} set.}
    \vspace{-0.6cm}
    \begin{center}
    \small{
        \begin{tabular}{l|c}
            \toprule
            \textbf{Method} & \textbf{PQ} \\
            \midrule
             Panoptic-PolarNet~\cite{zhou2021panoptic} & $54.1$ \\
             DS-Net~\cite{dsnet} & $55.9$ \\
             EfficientLPS~\cite{sirohi2021efficientlps} & $57.4$ \\
             GP-S3Net~\cite{razani2021gp} & $60.0$ \\
             SCAN~\cite{xu2022sparse} & $61.5$ \\
             Panoptic-PHNet~\cite{li2022panoptic} & $64.6$ \\
             \midrule
            \rowcolor{cyan!9}\textbf{\algorithmname~(Ours)} & $\mathbf{67.2}$ \\
            \bottomrule
        \end{tabular}
    }
    \end{center}
    \label{tab:sem_ps}
    \vspace{-0.3cm}
\end{table}

\begin{table}[!t]
    \caption{Quantitative results of \algorithmname~and SoTA LiDAR panoptic segmentation methods on nuScenes \textit{test} set.
    }
    \vspace{-0.6cm}
    \begin{center}
    \small{
        \begin{tabular}{l|c}
            \toprule
            \textbf{Method} & \textbf{PQ} \\
            \midrule
           EfficientLPS~\cite{sirohi2021efficientlps} & $62.4$ \\
            Panoptic-PolarNet~\cite{zhou2021panoptic} & $63.6$ \\
           
           SPVNAS~\cite{tang2020searching} + CenterPoint~\cite{yin2021center} & $72.2$ \\
           
           Cylinder3D++~\cite{zhu2021cylindrical} + CenterPoint~\cite{yin2021center} & $76.5$ \\
           
          AF2S3Net~\cite{af2s3net} + CenterPoint~\cite{yin2021center} & $76.8$ \\
            SPVCNN++~\cite{tang2020searching} & $79.1$ \\
            LidarMultiNet~\cite{lidarmultinet} & $81.4$ \\
            
            Panoptic-PHNet~\cite{li2022panoptic} & $\mathbf{81.5}$ \\
            
            \midrule
            \rowcolor{cyan!9}\textbf{\algorithmname~(Ours)} & $78.4$  \\
            \bottomrule
        \end{tabular}
    }
    \end{center}
    \label{tab:sem_nusc}
    \vspace{-0.3cm}
\end{table}

\noindent \textbf{Multi-Modal Fusion Baselines}. We take classical early fusion, PointPainting~\cite{pointpainting} and PointAugmenting~\cite{pointaugmenting} as multi-modal fusion baselines. Early fusion conducts input-level fusion and we select two early fusion variants, \ie, addition and concatenation of input signals. PointPainting appends the point cloud with the semantic segmentation scores while PointAugmenting fuses the point cloud with the image features of the segmentation branch.

\subsection{Comparative Study}

\noindent \textbf{Quantitative Results}. We summarize the performance of \algorithmname~and state-of-the-art LiDAR segmentation methods in Table~\ref{tab:sem_ss}-\ref{tab:sem_waymo}. For LiDAR semantic segmentation, our \algorithmname~outperforms the competitive 2DPASS~\cite{yan20222dpass} by \textbf{2.3} mIoU. For classes of bicycle, motorcycle, and other vehicles, \algorithmname~is at least 8 IoU higher than 2DPASS~\cite{yan20222dpass}. As to panoptic segmentation, \algorithmname~achieves 67.2 PQ, surpassing the rival Panoptic-PHNet~\cite{li2022panoptic} by \textbf{2.6} PQ. On the nuScenes benchmark, \algorithmname~obtains \textbf{83.5} mIoU on the LiDAR semantic segmentation task and outperforms the second place, \ie, LidarMultiNet~\cite{lidarmultinet}, by \textbf{2.1} mIoU. As for panoptic segmentation, our \algorithmname~achieves 78.4 PQ and is on par with competitive panoptic segmentation algorithms such as SPVCNN++. Encouraging results are also observed in the WOD val set. \algorithmname~obtains 69.6 mIoU and is \textbf{2.2} mIoU higher than SPVCNN\cite{tang2020searching}. The impressive experimental results strongly prove the effectiveness of the presented multi-modal fusion network.

\begin{table}[!t]
    \caption{Quantitative results of UniSeg and SoTA LiDAR semantic segmentation methods on the WOD \textit{val} set. Methods with * denote our implementations.
    }
    \vspace{-0.6cm}
    \begin{center}
    \small{
        \begin{tabular}{l|c}
            \toprule
            \textbf{Method} & \textbf{mIoU} \\
            \midrule
            Point Transformer*~\cite{point-transformer} & $63.3$ \\
            Cylinder3D*~\cite{zhu2021cylindrical} & $66.0$ \\
            SPVCNN*~\cite{tang2020searching} & $67.4$ \\
            \midrule
            \rowcolor{cyan!9}\textbf{\algorithmname~(Ours)} & $\mathbf{69.6}$  \\
            \bottomrule
        \end{tabular}
    }
    \end{center}
    \label{tab:sem_waymo}
    \vspace{-0.3cm}
\end{table}

\begin{table}
    \caption{The comparisons between efficiency (run-time) and accuracy (mIoU) on the SemanticKITTI \textit{val} set.
    }
    \vspace{-0.6cm}
    \begin{center}
    \small{
        \begin{tabular}{r|c|c| c}
            \toprule
            \textbf{Method} & \textbf{\#Param} & \textbf{Latency} & \textbf{mIoU}  \\
            \midrule
            Cylinder3D~\cite{zhu2021cylindrical}   & $56.3$M & $75.1$ms  &  $65.9$
            \\
            MinkowskiNet~\cite{2019Minkowski} & $21.7$M & $48.4$ms  &  $61.1$
            \\
            SPVCNN~\cite{tang2020searching} & $21.8$M & $52.4$ms &  $63.8$
            \\
            \midrule
            \rowcolor{cyan!9}\textbf{UniSeg $0.2$$\times$~(Ours)}  & $28.8$M & $84.6$ms & $\mathbf{67.0}$
            \\
            \rowcolor{cyan!9}\textbf{UniSeg $1.0$$\times$~(Ours)}   & $147.6$M & $145.0$ms  & $\mathbf{71.3}$
            \\
            \bottomrule
        \end{tabular}
    }
    \end{center}
    \label{tab:latency_kitti}
    \vspace{-0.3cm}
\end{table}

\begin{table}[!t]
    \caption{Comparison with different multi-modal feature fusion strategies on the SemanticKITTI \textit{val} set. Methods with * denote our implementations.
    }
    \vspace{-0.6cm}
    \begin{center}
    \small{
        \begin{tabular}{l|c|c}
            \toprule
            \textbf{Method} & \textbf{mIoU} & $\Delta$ \\
            \midrule
            Early Fusion Add (Baseline)~ & $70.1$ & $+0.0$
            \\\midrule
            Early Fusion Concat~ & $69.4$ & $-0.7$ \\
            
            PointPainting*~\cite{pointpainting} & $70.4$ & $+0.3$ \\
            PointAugmenting*~\cite{pointaugmenting} & 70.5 & $+0.4$ \\\midrule
            \rowcolor{cyan!9}\textbf{LMA~(Ours)} & $\mathbf{71.3}$ & $+1.2$ \\
            \bottomrule
        \end{tabular}
    }
    \end{center}
    \label{tab:fea_fuse}
    \vspace{-0.3cm}
\end{table}

\noindent \textbf{Comparisons of Efficiency and Accuracy}. We provide comparisons of efficiency and accuracy as shown in Table \ref{tab:latency_kitti}, our UniSeg\_0.2$\times$ achieves the best accuracy when the parameters and latency are comparable to other methods. Note that UniSeg\_0.2$\times$ is produced from the original UniSeg model by pruning 80\% channels for each layer. Besides, when increasing the parameters, the accuracy is further improved (UniSeg). All models are tested at NVIDIA A100 GPU.

\noindent \textbf{Is the Implementation Optimal?}
We would like to show that the implementation achieves the best performance after trials and errors. Specifically, \textbf{For the LMA module:} considering the calibration errors caused by the imperfect calibration matrices between the LiDAR and the camera. We have made several attempts to alleviate this issue (Table~\ref{tab:fea_fuse}). Firstly, we directly added or concatenated the image-point feature, and achieved +0.4 mIoU and -0.3 mIoU, respectively. Secondly, we adopt PointPainting~\cite{pointpainting} and PointAugmenting~\cite{pointaugmenting} to fuse feature, the improvement is 0.7 mIoU and 0.8 mIoU, respectively, but these fusion methods are sensitive to calibration errors. Thirdly, We tried the Self-attention operation. However, it suffers from the high computational cost introduced by the global-wise attention calculation. Lastly, we adopt the Deformable cross-attention in our method due to its efficiency and effectiveness. As shown in Table \ref{tab:fea_fuse}, the LMA module improved \textbf{1.6} mIoU and outperformed add, concatenate, PointPainting, and PointAugmenting by 1.2, 1.9, 0.9, and 0.8 in mIoU, respectively. 

\textbf{For LVA module:} We explore how to leverage the advantages of different modality data. Firstly, we conduct the baseline method, i.e., it transfers all modality data into the point-view and then directly adds or concatenates them, the performance is 70.4 mIoU and 70.5 mIoU, respectively. Secondly, we tried self-attention for feature fusion but could not achieve improvement. Lastly, we design the LVA module to adaptively fuse the different modality data based on the learned attention weights. As shown in Table \ref{tab:view_fuse}, the improvement is \textbf{0.9} mIoU compared to the direct addition and concatenation.

\subsection{Ablation Study}

We perform an ablation study to verify the effect of each modality/view and different cross-view fusion variants on the final performance. The following experiments are conducted in the SemanticKITTI validation set.

\begin{table}[!t]
\caption{Influence of different modalities and views.}
\label{each_modal_table}
\centering
\vspace{-0.3cm}
\small{
\begin{tabular}{l|c|c|c|c}
\toprule
\textbf{Voxel} & \textbf{Point} & \textbf{Range image} & \textbf{RGB Image} & \textbf{mIoU} \\
\midrule
$\checkmark$ &         &         &         &   $68.4$ \\
        & $\checkmark$ &         &         &  $13.7$ \\
        &         & $\checkmark$ &         &  $55.8$ \\
$\checkmark$ &         &         & $\checkmark$ &  $69.7$ \\
 &         &    $\checkmark$     & $\checkmark$ &  $58.1$ \\

$\checkmark$ & $\checkmark$ &         &         &  $68.5$ \\
$\checkmark$ & $\checkmark$ & $\checkmark$ &         &  $69.7$
\\\midrule
\rowcolor{cyan!9}$\checkmark$ & $\checkmark$ & $\checkmark$ & $\checkmark$ &  $\mathbf{71.3}$ \\
\bottomrule
\end{tabular}
}
\end{table}

\begin{table}
    \caption{Robustness on the SemanticKITTI \textit{val} set. The symbol * denotes calibration matrices with noises.
    }
    \vspace{-0.6cm}
    \begin{center}
    \small{
        \begin{tabular}{r|c|c|c|c}
            \toprule
            \textbf{Method}  & Add    & \cellcolor{cyan!9}LMA &   Add* &    \cellcolor{cyan!9}LMA*  \\
            \midrule
            \textbf{mIoU} &   $70.1$ & \cellcolor{cyan!9}$\mathbf{71.3}$ & $68.5$ & \cellcolor{cyan!9}$71.0$   \\
            \bottomrule
        \end{tabular}
    }
    \end{center}
    \label{tab:Robustness}
    \vspace{-0.3cm}
\end{table}

\begin{table}[!t]
    \caption{Comparisons among cross-view fusion strategies.
    }
    \vspace{-0.6cm}
    \begin{center}
    \small{
        \begin{tabular}{c|c|c}
            \toprule
            \textbf{Method} & \textbf{mIoU} & $\Delta$
            \\
            \midrule
            Add (Baseline)~ & $70.4$ & $+0.0$
            \\\midrule
            Concat~ & $70.5$ & $+0.1$
            \\
            Self-Attention~ & $70.4$ & $+0.0$
            \\
            \midrule
            \rowcolor{cyan!9}\textbf{LVA~} & $\mathbf{71.3}$ & $+0.9$ \\
            \bottomrule
        \end{tabular}
    }
    \end{center}
    \label{tab:view_fuse}
    \vspace{-0.6cm}
\end{table}

\begin{figure}
\centering
\includegraphics[width=0.9\linewidth]{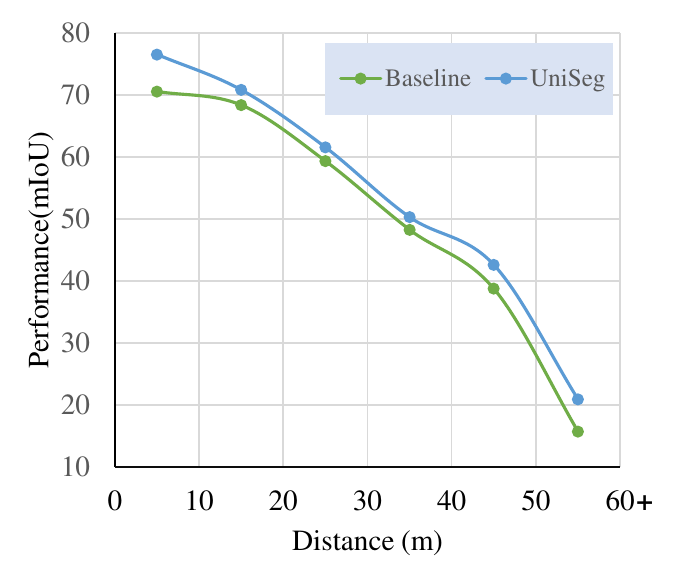}
\vskip -0.4cm
\caption{Comparison between the single-modal baseline and UniSeg with different distances on SemanticKITTI.}
\centering
\vspace{-8pt}
\label{fig:distance}
\end{figure}

\noindent\textbf{Effect of Each Modality}. We summarize the influence of each modality as well as their combinations on the final performance in Table~\ref{each_modal_table}. From the first three rows, we can see that the voxel branch exhibits much better performance than the other two representations, showing the indispensable role of the voxel representation. Fusing three views of the point cloud with images yield the best performance, demonstrating the value of every single modality on the segmentation results. Besides, our \algorithmname~also outperforms the single-modal baseline in different distances (Fig. \ref{fig:distance}). Obviously, the baseline degrades at a long distance due to more sparsity. And \algorithmname~consistently outperforms the uni-modal baseline, strongly demonstrating the value of the multi-modal representation.

\noindent\textbf{Fusion Strategies}. We compare our proposed LMA module with other fusion strategies as shown in Table \ref{tab:fea_fuse}, it brings a larger improvement than other methods and outperforms 1.2 mIoU than baseline. Notably, when we used UniSeg\_0.2$\times$ to compare the LMA module with PointPainting, the LMA module was \textbf{1.5} mIoU higher than PointPainting, which directly demonstrates the benefits of the LMA module. With the help of the LVA module, the point-, voxel-, and range-view features are more effectively fused compared with other fusion methods as shown in Table~\ref{tab:view_fuse}.

\noindent\textbf{Robustness to calibration error}. We add Gaussian noise to the calibration matrices to evaluate the robustness. As shown in Table \ref{tab:Robustness}, {\algorithmname} drops \textbf{0.3} mIoU while the addition operation drops \textbf{1.6} mIoU, indicating the LMA module is more tolerant to calibration noise.

\begin{figure*}[h]
    \begin{center}
    \includegraphics[width=0.861\linewidth]{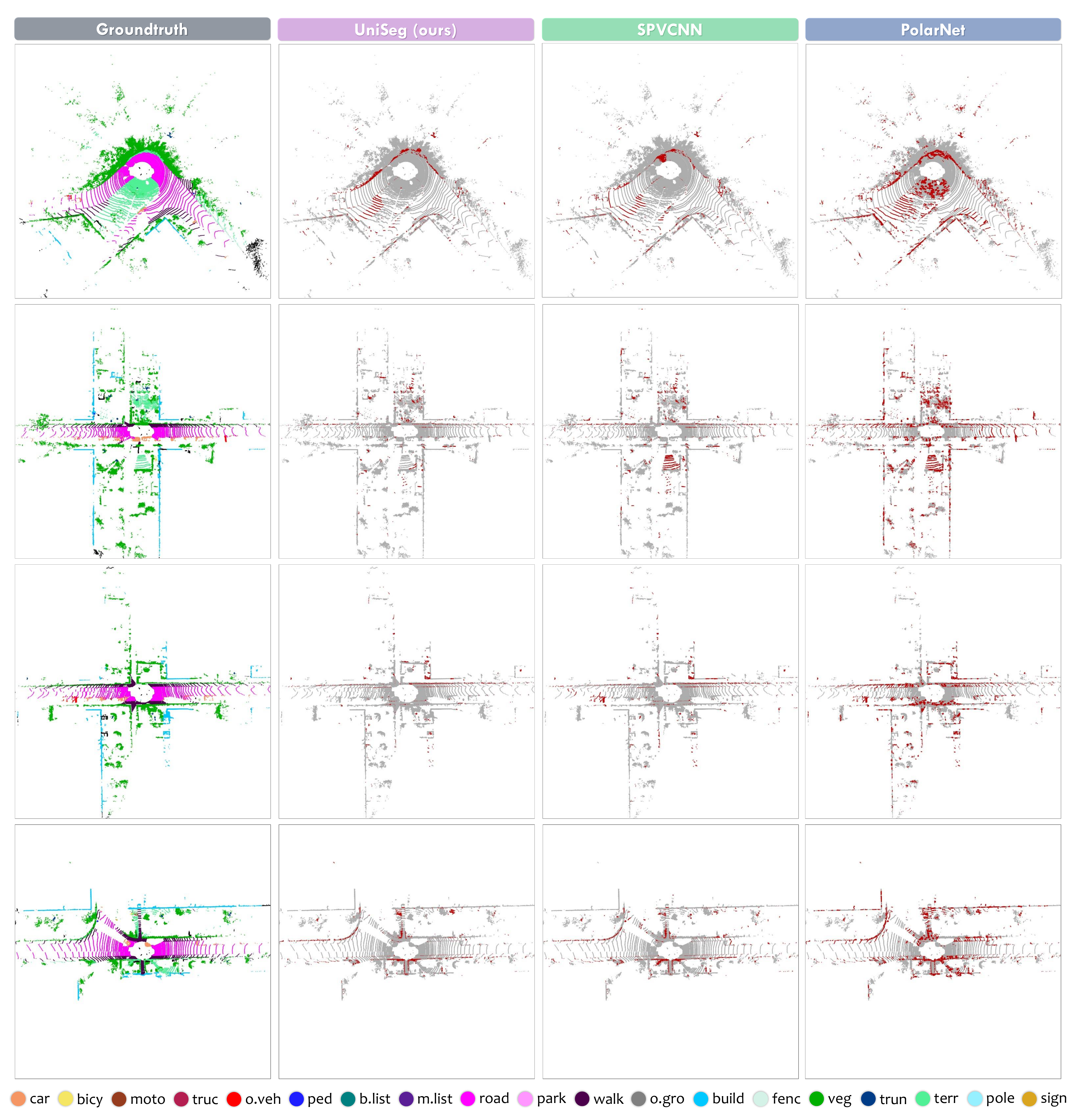}
    \end{center}
    \vspace{-0.5cm}
    \caption{\textbf{Qualitative comparisons} with SPVCNN~\cite{tan2019efficientnet} and PolarNet~\cite{panopticpolarnet} through \textbf{error maps}. To highlight the differences, the \textbf{\textcolor{correct}{correct}} / \textbf{\textcolor{incorrect}{incorrect}} predictions are painted in \textbf{\textcolor{correct}{gray}} / \textbf{\textcolor{incorrect}{red}}, respectively. Each scene is visualized from the LiDAR bird's eye view and covers a region of size $50$m by $50$m, centered around the ego-vehicle. Best viewed in colors.}
    \label{figure:qualitative_supp}
    \vspace{0pt}
\end{figure*}

\subsection{Qualitative Results}

We provide qualitative comparisons with SPVCNN~\cite{tan2019efficientnet} and PolarNet~\cite{panopticpolarnet} through error maps in Fig.~\ref{figure:qualitative_supp}. Upon examining the results, it becomes evident that our approach demonstrates superior performance while maintaining minimal segmentation errors across each sampled frame.

\section{Conclusion}
\label{conclusion}
We propose a unified multi-modal LiDAR segmentation network, dubbed \algorithmname, that makes the first attempt to take RGB images and three views of the point cloud as input, and performs semantic and panoptic segmentation simultaneously. To fully leverage the information of different modalities data, we present the cross-Modal Association module (LMA) and the Learnable cross-View Association module (LVA). Equipped with LMA and LVA, \algorithmname~achieves compelling performance in three popular LiDAR segmentation benchmarks and ranks 1st in two open challenges.

\vspace{5pt}

\textbf{Acknowledgements.} This work is supported by the Science and Technology Commission of Shanghai Municipality (grant No. 22DZ1100102).

\section*{Appendix}
In this file, we supplement additional materials to support our findings, observations, and experimental results. 
Specifically, this file is organized as follows:
\begin{itemize}
    \item \cref{sec:openpcseg} provides additional information on the OpenPCSeg codebase and summarizes the reproduced and reported performance.
    \item \cref{sec:implementation} elaborates on additional implementation details of the proposed methods and the experiments. 
    \item \cref{sec:quan} supplements additional quantitative results, including class-wise IoU scores and PQ scores for our comparative study and ablation study.
    \item \cref{sec:qual} attaches additional qualitative results.
\end{itemize}

\section{Additional Information of OpenPCSeg}
\label{sec:openpcseg}

The OpenPCSeg codebase supports tasks of LiDAR semantic segmentation and LiDAR panoptic segmentation. It includes range-image-based, voxel-based, fusion-based, point-based and BEV-based algorithms, as well as recent 3D data augmentation techniques. Range-image-based methods include SqueezeSeg~\cite{wu2018squeezeseg}, SqueezeSegV2~\cite{wu2019squeezesegv2}, RangeNet++~\cite{milioto2019rangenet++}, FIDNet~\cite{zhao2021fidnet}, CENet~\cite{cheng2022cenet} and SalsaNext~\cite{cortinhal2020salsanext}. Voxel-based algorithms have MinkowskiNet~\cite{choy20194d}, Cylinder3D~\cite{zhu2021cylindrical}, and
DS-Net~\cite{dsnet}. Fusion-based algorithms include RPVNet~\cite{rpvnet} and SPVCNN~\cite{tang2020searching}. Point-based algorithms contain PointTransformer~\cite{point-transformer}. BEV-based algorithms including PolarNet~\cite{zhang2020polarnet}, and Panoptic-PolarNet~\cite{panopticpolarnet}. We also have three useful data augmentation algorithms, LaserMix~\cite{lasermix}, PolarMix~\cite{xiao2022polarmix}, Mix3D~\cite{nekrasov2021mix3d}. A summary of supported features compared to the existing codebase is provided in Table~\ref{tab:codebase-compare}. OpenPCSeg supports more datasets and more features than other codebases. A detailed comparison between the reproduced and reported performance of different algorithms is summarized in Table~\ref{tab:openpcseg_semseg_sup}. Besides, we provide  MinkowskiNet~\cite{choy20194d} and SPVCNN~\cite{tan2019efficientnet} variants are shown in Table~\ref{tab:variants}. More popular LiDAR segmentation algorithms, such as Panoptic-PHNet~\cite{li2022panoptic} and LidarMultiNet~\cite{lidarmultinet}, will be added to this codebase in the future. 

We elaborate on more details of the benchmarked models, techniques, and datasets as follows.

\begin{table*}[t]
    \centering
    \renewcommand\arraystretch{1.4}
    \caption{Supported features of existing LiDAR segmentation codebases. ``~\checkmark~" / ``~\ding{53}~" denotes a supported / not supported feature. Symbol ``$\triangle$" denotes a feature that is to be supported in future updates.}
    \vskip -0.2cm
    \label{tab:codebase-compare}
    \footnotesize
    \scalebox{0.92}{
    \begin{tabular}{c|c|p{63pt}<{\centering}|p{63pt}<{\centering}|p{63pt}<{\centering}|p{63pt}<{\centering}|p{63pt}<{\centering}}
    \toprule
    \textbf{Type} & \textbf{Feature} & \textbf{MMDetection3D}\footnote{\url{https://github.com/open-mmlab/mmdetection3d}.} & \textbf{3D-SemSeg}\footnote{\url{https://github.com/danielmohansahu/benchmarking-3d-semantic-segmentation}.} & \textbf{lidarseg3d}\footnote{\url{https://github.com/jialeli1/lidarseg3d}.} & \textbf{Open3D-ML}\footnote{\url{https://github.com/isl-org/Open3D-ML\#model-zoo}.} & \cellcolor{cyan!9}\textbf{OpenPCSeg~(Ours)}
    \\\midrule\midrule
    \multirow{3}*{\textbf{Task}} 
    & Semantic Segmentation & \checkmark   & \checkmark & \checkmark & \checkmark & \cellcolor{cyan!9}\checkmark
    \\
    & Panoptic Segmentation &  \ding{53}  & \ding{53} & \ding{53} & \ding{53} & \cellcolor{cyan!9}\checkmark  
    \\
    & 4D Panoptic Segmentation &  \ding{53}  & \ding{53} & \ding{53} & \ding{53} & \cellcolor{cyan!9}\checkmark 
    \\\midrule
    \multirow{4}*{\textbf{Dataset}} & SemanticKITTI &  \checkmark & \checkmark & \checkmark & \checkmark & \cellcolor{cyan!9}\checkmark
    \\
    & nuScenes &    \ding{53}  & \checkmark & \checkmark   & \ding{53}          & \cellcolor{cyan!9}\checkmark
    \\ 
    & Waymo Open &  \ding{53} &  \ding{53} & \ding{53}    & \ding{53}          & \cellcolor{cyan!9}\checkmark
    \\
    & ScribbleKITTI & \ding{53} & \ding{53} & \ding{53}    & \ding{53}            & \cellcolor{cyan!9}\checkmark
    \\\midrule
    \multirow{25}*{\textbf{Model}}
    & SqueezeSeg           &  \ding{53}  & \ding{53} & \ding{53}    & \ding{53}       & \cellcolor{cyan!9}\checkmark \\
    & SqueezeSegV2         &   \ding{53} & \ding{53} & \ding{53}     & \ding{53}       & \cellcolor{cyan!9}\checkmark
    \\
    & RangeNet++           &  \ding{53}  & \ding{53} & \ding{53} & \ding{53}        & \cellcolor{cyan!9}\checkmark
    \\
    & SalsaNext            &  \ding{53}  & \checkmark  & \ding{53}    & \ding{53}       & \cellcolor{cyan!9}\checkmark \\
    & FIDNet               & \ding{53}   & \ding{53}  & \ding{53}     & \ding{53}        & \cellcolor{cyan!9}\checkmark    \\
    & CENet                &  \ding{53}  & \ding{53} & \ding{53}     & \ding{53}          & \cellcolor{cyan!9}\checkmark\\
    & PolarNet             & \ding{53}   & \ding{53} & \ding{53}      & \ding{53}         & \cellcolor{cyan!9}\checkmark\\
    & Panoptic-PolarNet    & \ding{53}   & \ding{53} & \ding{53}     & \ding{53}          & \cellcolor{cyan!9}\checkmark\\
    & RandLA-Net          & \ding{53}   & \ding{53} & \ding{53}     & \checkmark          & \cellcolor{cyan!9}\ding{53}
    \\
    & KPConv               & \ding{53}   & \ding{53} & \ding{53}     & \checkmark          & \cellcolor{cyan!9}\ding{53}
    \\
    & SparseConvUnet         & \ding{53}   & \ding{53} & \ding{53}     & \checkmark          & \cellcolor{cyan!9}\ding{53}
    \\
    & PointTransformer        & \ding{53}   & \ding{53} & \ding{53}     & \checkmark          & \cellcolor{cyan!9}$\triangle$
    \\
    & PointNet++               & \checkmark   & \ding{53} & \ding{53}     & \ding{53}          & \cellcolor{cyan!9}\ding{53}
    \\
    & PAConv                    & \checkmark   & \ding{53} & \ding{53}     & \ding{53}          & \cellcolor{cyan!9}\ding{53}
    \\
    & DGCNN                     & \checkmark   & \ding{53} & \ding{53}     & \ding{53}          & \cellcolor{cyan!9}$\triangle$
    \\
    & MinkowskiNet         &  \ding{53}  & \ding{53} & \ding{53}       & \ding{53}       & \cellcolor{cyan!9}\checkmark
    \\
    & Cylinder3D           &  \ding{53}  & \checkmark  & \ding{53}       & \ding{53}     & \cellcolor{cyan!9}\checkmark
    \\
    & DS-Net               &  \ding{53}  & \ding{53} & \ding{53}         & \ding{53}     & \cellcolor{cyan!9}\checkmark
    \\
    & 4D-DS-Net               &  \ding{53}  & \ding{53} & \ding{53}         & \ding{53}     & \cellcolor{cyan!9}\checkmark
    \\
    & RPVNet               &  \ding{53}  & \ding{53} & \ding{53}        & \ding{53}     & \cellcolor{cyan!9}\checkmark
    \\
    & SPVCNN               & \ding{53}   & \ding{53} & \ding{53}        & \ding{53}      & \cellcolor{cyan!9}\checkmark
    \\
    & 2DPASS               &  \ding{53} &  \checkmark & \ding{53}       & \ding{53}      & \cellcolor{cyan!9}$\triangle$
    \\
    & COARSE3D             &  \ding{53} &  \checkmark & \ding{53}        & \ding{53}     & \cellcolor{cyan!9}$\triangle$
    \\
    & SDSeg3D              &  \ding{53} &  \ding{53} & \checkmark        & \ding{53}   & \cellcolor{cyan!9}\ding{53}
    \\
    & MSeg3D               &  \ding{53} &  \ding{53} & $\triangle$        & \ding{53}    & \cellcolor{cyan!9}\ding{53}
    \\\midrule
    \multirow{3}*{\textbf{Augmentation}} 
    & Mix3D                &  \ding{53}  & \ding{53} & \ding{53} & \ding{53} & \cellcolor{cyan!9}\checkmark
    \\
    & LaserMix             &  \ding{53}   & \ding{53} & \ding{53} & \ding{53} & \cellcolor{cyan!9}\checkmark
    \\
    & PolarMix             &  \ding{53}   & \ding{53} & \ding{53} & \ding{53} & \cellcolor{cyan!9}\checkmark
    \\\midrule\midrule
    & \textbf{\# Supported Features} & $5$ & $7$ & $5$ & $6$ & \cellcolor{cyan!9}$28$
    \\\bottomrule
    \end{tabular}}
\vspace{0.1cm}
\end{table*}

\begin{table*}[t]
\caption{Comparisons between the reproduced performance in the \textbf{OpenPCSeg codebase} (mIoU-rep, PQ-rep) and reported performance from the original papers (mIoU-ori, PQ-ori). We benchmark various popular LiDAR semantic segmentation methods and LiDAR panoptic segmentation methods on the validation sets of SemanticKITTI~\cite{behley2019semantickitti} and nuScenes~\cite{caesar2020nuscenes}. Note that we only report range-view methods with sizes $64 \times 2048$ and $32 \times 1920$ for SemanticKITTI and nuScenes, respectively. }
\label{tab:openpcseg_semseg_sup}
\vskip -0.2cm
\centering\scalebox{0.85}{
\begin{tabular}{r|c|c|c|c|c|c|c|c|c}
\toprule
\multirow{2}*{\textbf{Model}} & \multirow{2}*{\textbf{Type}} & \multicolumn{4}{c|}{\textbf{SemanticKITTI}} & \multicolumn{4}{c}{\textbf{nuScenes}} \\
~ & ~ & mIoU-ori & \cellcolor{cyan!9}mIoU-rep & PQ-ori & \cellcolor{cyan!9}PQ-rep & mIoU-ori & \cellcolor{cyan!9}mIoU-rep & PQ-ori & \cellcolor{cyan!9}PQ-rep \\
\midrule\midrule
Mix3D~\cite{nekrasov2021mix3d} &\multirow{3}*{Aug}& -- & \cellcolor{cyan!9}-- & -- & \cellcolor{cyan!9}-- & -- & \cellcolor{cyan!9}-- & -- & \cellcolor{cyan!9}--
\\
LaserMix~\cite{lasermix} &~& -- & \cellcolor{cyan!9}--  & --  & \cellcolor{cyan!9}--  & -- & \cellcolor{cyan!9}--  & -- & \cellcolor{cyan!9}--
\\
PolarMix~\cite{xiao2022polarmix} &~& -- & \cellcolor{cyan!9}-- & --  & \cellcolor{cyan!9}-- & -- & \cellcolor{cyan!9}-- & -- & \cellcolor{cyan!9}--
\\
\midrule
SqueezeSeg~\cite{wu2018squeezeseg} &\multirow{8}*{Range}& $31.6$ & \cellcolor{cyan!9}$33.0_{\textcolor{blue}{(+1.4)}}$ & -- & \cellcolor{cyan!9}--  & --  & \cellcolor{cyan!9}--  & --  & \cellcolor{cyan!9}-- 
\\
SqueezeSegV2~\cite{wu2019squeezesegv2} &~& $41.3$ & \cellcolor{cyan!9}$44.5_{\textcolor{blue}{(+3.2)}}$ & -- & \cellcolor{cyan!9}--  & --  & \cellcolor{cyan!9}--  & --  & \cellcolor{cyan!9}-- 
\\
RangeNet$_{21}$~\cite{milioto2019rangenet++} &~& $47.2$ & \cellcolor{cyan!9}$49.8_{\textcolor{blue}{(+2.6)}}$ & -- & \cellcolor{cyan!9}--  & --  & \cellcolor{cyan!9}--  & --  & \cellcolor{cyan!9}-- 
\\
RangeNet$_{53}$~\cite{milioto2019rangenet++} &~& $50.3$ & \cellcolor{cyan!9}$53.3_{\textcolor{blue}{(+3.0)}}$ & -- & \cellcolor{cyan!9}--  & --  & \cellcolor{cyan!9}--  & --  & \cellcolor{cyan!9}--
\\
RangeNet$_{53}$++~\cite{milioto2019rangenet++} &~& $52.2$ & \cellcolor{cyan!9}$54.0_{\textcolor{blue}{(+1.8)}}$  & -- & \cellcolor{cyan!9}--  & --  & \cellcolor{cyan!9}$65.8$ & --  & \cellcolor{cyan!9}--
\\
SalsaNext~\cite{cortinhal2020salsanext} & ~ &  $55.8$ & \cellcolor{cyan!9}$58.2_{\textcolor{blue}{(+2.4)}}$  & -- & \cellcolor{cyan!9}--  & --  & \cellcolor{cyan!9}$68.1$ & --  & \cellcolor{cyan!9}--
\\
FIDNet~\cite{zhao2021fidnet}  &  & $58.8$ & \cellcolor{cyan!9}$60.4_{\textcolor{blue}{(+2.6)}}$ & --  & \cellcolor{cyan!9}-- & --  & \cellcolor{cyan!9}71.8 & --  & \cellcolor{cyan!9}--
\\
CENet~\cite{cheng2022cenet} & ~ & $62.6$ & \cellcolor{cyan!9}$63.7_{\textcolor{blue}{(+1.1)}}$ & -- & \cellcolor{cyan!9}--  & -- & \cellcolor{cyan!9}$73.4$ & -- & \cellcolor{cyan!9}--
\\\midrule
PolarNet~\cite{zhang2020polarnet} & \multirow{2}*{BEV} & $57.2$ & \cellcolor{cyan!9}$58.3_{\textcolor{blue}{(+1.1)}}$  & --  & \cellcolor{cyan!9}-- & -- & \cellcolor{cyan!9}$71.4$ & --  & \cellcolor{cyan!9}--
\\
Panoptic-PolarNet~\cite{panopticpolarnet} &~& -- & \cellcolor{cyan!9}-- & $59.1$ & \cellcolor{cyan!9}$59.5_{\textcolor{blue}{(+0.4)}}$ & -- & \cellcolor{cyan!9}-- & $67.7$ & \cellcolor{cyan!9}$67.8_{\textcolor{blue}{(+0.1)}}$
\\\midrule
MinkowskiNet~\cite{choy20194d} & \multirow{3}*{Voxel} &$61.1$  & \cellcolor{cyan!9}$68.8_{\textcolor{blue}{(+7.7)}}$ & --  & \cellcolor{cyan!9}-- &-- & \cellcolor{cyan!9}$73.2$ & --  & \cellcolor{cyan!9}--
\\
Cylinder3D~\cite{zhu2021cylindrical} & ~ & $65.9$ & \cellcolor{cyan!9}$66.9_{\textcolor{blue}{(+1.0)}}$  & --  & \cellcolor{cyan!9}-- & $76.1$ & \cellcolor{cyan!9}$76.2_{\textcolor{blue}{(+0.1)}}$ & --  & \cellcolor{cyan!9}--
\\
DS-Net~\cite{dsnet} &~& -- & \cellcolor{cyan!9}-- & $57.7$ & \cellcolor{cyan!9}$58.0_{\textcolor{blue}{(+0.3)}}$ & -- & \cellcolor{cyan!9}-- & $42.5$  & \cellcolor{cyan!9}$61.0_{\textcolor{blue}{(+18.5)}}$
\\
\midrule
RPVNet~\cite{rpvnet} & \multirow{3}*{Fusion} & $68.3$ & \cellcolor{cyan!9}$68.8_{\textcolor{blue}{(+0.5)}}$ & --  & \cellcolor{cyan!9}-- & $77.6$ & \cellcolor{cyan!9}$77.6_{\textcolor{blue}{(+0.0)}}$ & --  & \cellcolor{cyan!9}--
\\
SPVCNN~\cite{tang2020searching} & ~ &  $63.8$ & \cellcolor{cyan!9}$68.7_{\textcolor{blue}{(+3.9)}}$ & --  & \cellcolor{cyan!9}--  & --  & \cellcolor{cyan!9}$74.8$ & --  & \cellcolor{cyan!9}--
\\

\bottomrule
\end{tabular}
}
\vspace{0.2cm}
\end{table*}

\subsection{Supported LiDAR Segmentation Model}

\subsubsection{Range View}

\begin{itemize}
    \item SqueezeSeg~\cite{wu2018squeezeseg}: a classic 3D segmentor which can be trained end-to-end, proposed in 2017.
    \item SqueezeSegV2~\cite{wu2019squeezesegv2}: an improvement over SqueezeSeg by the Context Aggregation Module (CAM) to mitigate the impact of dropout noise, proposed in 2018. 
    \item RangeNet++~\cite{milioto2019rangenet++}: a classic and widely used range view LiDAR semantic segmentation method which equips with GPU-enabled post-processing, proposed in 2019. 
    \item SalsaNext~\cite{cortinhal2020salsanext}: a range-view solution for LiDAR semantic segmentation task which brings a Bayesian treatment to compute the \textit{epistemic} and \textit{aleatoric} uncertainties for each point, proposed in 2020.
    \item FIDNet~\cite{zhao2021fidnet}: a 3D segmentor with an improved post-processing method (NLA) over RangeNet++ and equips with an FID module for upsampling, proposed in 2021.
    \item CENet~\cite{cheng2022cenet} a powerful range view method embedding multiple auxiliary segmentation heads for LiDAR segmentation task, proposed in 2022.
    \item COARSE3D~\cite{li2022coarse3d}: a weakly supervised LiDAR semantic segmentation framework with a compact class-prototype contrastive learning scheme, proposed in 2022.
\end{itemize}

\subsubsection{Bird's Eye View}

\begin{itemize}
    \item PolarNet~\cite{zhang2020polarnet}: a classic 3D segmentor which quantizing points into polar bird’s-eye-view (BEV) grids, proposed in 2020.
    \item Panoptic-PolarNet~\cite{panopticpolarnet}: learn both semantic segmentation and class-agnostic instance clustering in a single network using a BEV representation to perform LiDAR panoptic segmentation task, proposed in 2021.
\end{itemize}

\subsubsection{Point View}

\begin{itemize}
    \item PointTransformer~\cite{point-transformer}: a powerful 3D network that is constructed with the Transformer architecture~\cite{vaswani2017attention}, proposed in 2021.
    \item DGCNN~\cite{wang2019dynamic}: a classic and widely used segmentation and classification method constructed by using EdgeConv, proposed in 2018.
\end{itemize}

\begin{table*}[t]
\caption{Comparisons among the variants of MinkowskiNet\cite{choy20194d} and SPVCNN\cite{tang2020searching} in the \textbf{OpenPCSeg codebase}. Results are on the validation sets of SemanticKITTI~\cite{behley2019semantickitti}, nuScenes~\cite{caesar2020nuscenes} and Waymo Open~\cite{waymo}. Symbol \textit{mk} denotes the number of layers of the network; Symbol \textit{cr} is the channel expansion rate. Note that the default setting of \textit{mk} and \textit{cr} are $18$ and $1.0$, respectively, for MinkowskiNet\cite{choy20194d} and SPVCNN\cite{tang2020searching}. }
\label{tab:variants}
\vskip -0.2cm
\centering\scalebox{0.85}{
\begin{tabular}{r|c|c|r|c|c|c|c|c|c}
\toprule
\multirow{2}*{\textbf{Model}} & \multirow{2}*{\textbf{Variant}} & \multirow{2}*{\textbf{Type}} & \multirow{2}*{\textbf{\#Param}} & \multicolumn{2}{c|}{\textbf{SemanticKITTI}} & \multicolumn{2}{c|}{\textbf{nuScenes}} & \multicolumn{2}{c}{\textbf{Waymo Open}} \\
~& ~ &~& & mIoU-ori & \cellcolor{cyan!9}mIoU-rep & mIoU-ori & \cellcolor{cyan!9}mIoU-rep & mIoU-ori & \cellcolor{cyan!9}mIoU-rep
\\\midrule\midrule
MinkowskiNet~\cite{choy20194d} & mk$18$cr$0.5$ & \multirow{4}*{Voxel} &  $5.5$ M & $58.9$  & \cellcolor{cyan!9}$68.7_{\textcolor{blue}{(+9.8)}}$ & -- & \cellcolor{cyan!9}-- & -- & \cellcolor{cyan!9}--
\\
MinkowskiNet~\cite{choy20194d} & mk$18$cr$1.0$ & ~ & $21.7$ M  &  $61.1$ & \cellcolor{cyan!9}$68.8_{\textcolor{blue}{(+7.7)}}$ & -- & \cellcolor{cyan!9}$73.2$ & -- & \cellcolor{cyan!9}$66.7$
\\
MinkowskiNet~\cite{choy20194d} & mk$34$cr$1.0$ &~& $37.9$ M  &  --       & \cellcolor{cyan!9}$70.1$  & -- & \cellcolor{cyan!9}$75.7$ & -- & \cellcolor{cyan!9}--
\\
MinkowskiNet~\cite{choy20194d} & mk$34$cr$1.6$ &~& $96.5$ M   &  --       & \cellcolor{cyan!9}$70.1$  & -- & \cellcolor{cyan!9}$76.2$ & -- & \cellcolor{cyan!9}$68.2$  \\
\midrule
SPVCNN~\cite{tang2020searching} & mk$18$cr$0.5$ & \multirow{4}*{Fusion} &  $5.5$ M & $60.7$  & \cellcolor{cyan!9}$68.7_{\textcolor{blue}{(+8.0)}}$ & -- & \cellcolor{cyan!9}-- & -- & \cellcolor{cyan!9}--
\\
SPVCNN~\cite{tang2020searching} & mk$18$cr$1.0$ & ~& $21.8$ M  &  $63.8$       & \cellcolor{cyan!9}$67.6_{\textcolor{blue}{(+3.8)}}$ & -- & \cellcolor{cyan!9}$74.8$ & -- & \cellcolor{cyan!9}$66.8$
\\
SPVCNN~\cite{tang2020searching} & mk$34$cr$1.0$ &~& $37.9$ M  &  --       & \cellcolor{cyan!9}$69.0$  & -- & \cellcolor{cyan!9}$76.1$ & -- & \cellcolor{cyan!9}--
\\
SPVCNN~\cite{tang2020searching} & mk$34$cr$1.6$ &~& $96.7$ M   &  --       & \cellcolor{cyan!9}$68.4$  & -- & \cellcolor{cyan!9}$76.8$ & -- & \cellcolor{cyan!9}$68.6$
\\\bottomrule
\end{tabular}
}
\end{table*}

\subsubsection{Voxel \& Cylinder}

\begin{itemize}
    \item MinkowskiNet~\cite{choy20194d}: a classic and widely used LiDAR segmentation method, proposed in 2019.
    \item Cylinder3D~\cite{zhu2021cylindrical}: a cylindrical and asymmetrical 3D convolution network for LiDAR semantic segmentation, proposed in 2021.
    \item DS-Net~\cite{dsnet}: adopts consensus-driven fusion module and the dynamic shifting module for LiDAR panoptic segmentation, proposed in 2021.
    \item 4D-DS-Net~\cite{hongunified}: an extensive network of DS-Net to perform 4D panoptic LiDAR segmentation via temporally unified instance clustering on the aligned adjacent LiDAR frames, proposed in 2022.
\end{itemize}

\subsubsection{Fusion}
\begin{itemize}
    \item SPVCNN~\cite{tang2020searching}: a powerful 3D segmentor adopt point-voxel fusion, proposed in 2020.
    \item RPVNet~\cite{rpvnet}: a multi-view LiDAR semantic segmentation method which includes range-point-voxel fusion, proposed in 2021.
    \item 2DPASS~\cite{xu2022sparse}: a new framework for LiDAR semantic segmentation via 2D prior-related knowledge distillation, proposed in 2022.
\end{itemize}

\subsection{Supported Data Augmentation Technique}
\begin{itemize}
    \item Mix3D~\cite{nekrasov2021mix3d}: a data augmentation technique for segmenting large-scale 3D scenes which build new training samples by mixing two augmented scenes, proposed in 2021.

    \item PolarMix~\cite{xiao2022polarmix}: a data augmentation technique that cuts, edits, and mixes point clouds along the scanning direction from two scenes, proposed in 2022.

    \item LaserMix~\cite{lasermix}: a powerful data augmentation technique that intertwines laser beams from different LiDAR scans, proposed in 2022.

\end{itemize}

\subsection{Supported LiDAR Segmentation Dataset}
\begin{itemize}
    \item SemanticKITTI~\cite{behley2019semantickitti}: a large-scale outdoor dataset for semantic scene understanding of LiDAR sequences collected from the 64-beam scan sensor, proposed in 2019.

    \item nuScenes~\cite{panoptic-nuscenes,caesar2020nuscenes}: a large-scale benchmark with support for various tasks, including camera images and LiDAR scans, and the point clouds are collected from the 32-beam scan sensor, proposed in 2020.

    \item Waymo Open~\cite{waymo}: A large-scale outdoor dataset consisting of well-synchronized and calibrated high-quality LiDAR and camera data, and the point clouds are collected from the 64-beam scan sensor, proposed in 2020.

    \item ScribbleKITTI~\cite{2022ScribbleKITTI}: is a recent variant of the SemanticKITTI dataset, which contains the same number of scans but is annotated with line scribbles (approximately 8.06\% valid semantic labels) rather than dense annotation, proposed in 2022.

\end{itemize}

\begin{table*}[ht]
\caption{Quantitative results of \algorithmname~and state-of-the-art \textbf{LiDAR semantic segmentation} methods on the \textit{test} set of \textbf{SemanticKITTI}~\cite{behley2019semantickitti}.}
\vskip -0.2cm
\label{tab:sem_ss_sup}
\centering
\begin{adjustbox}{width=\textwidth}
\begin{tabular}{r|c|c|c|c|c|c|c|c|c|c|c|c|c|c|c|c|c|c|c|c}
\toprule
\textbf{Model} & \rotatebox{90}{\textbf{mIoU}} & \rotatebox{90}{car} &  \rotatebox{90}{bicycle} & \rotatebox{90}{motorcycle} & \rotatebox{90}{truck} & \rotatebox{90}{other-vehicle} & \rotatebox{90}{person} & \rotatebox{90}{bicyclist} & \rotatebox{90}{motorcyclist} & \rotatebox{90}{road} & \rotatebox{90}{parking} & \rotatebox{90}{sidewalk} & \rotatebox{90}{other-ground} &
\rotatebox{90}{building} & \rotatebox{90}{fence} & \rotatebox{90}{vegetation} & \rotatebox{90}{trunk} & \rotatebox{90}{terrain} & \rotatebox{90}{pole} & \rotatebox{90}{traffic} \\
\midrule
\midrule
PointNet~\cite{qi2017pointnet} & $14.6$ & $46.3$ & $1.3$ & $0.3$ & $0.1$ & $0.8$ & $0.2$ & $0.2$ & $0.0$ & $61.6$ & $15.8$ & $35.7$ & $1.4$ & $41.4$ & $12.9$ & $31.0$ & $4.6$ & $17.6$ & $2.4$ & $3.7$
\\

PointNet++~\cite{qi2017pointnet++} & $20.1$ & $53.7$ & $1.9$ & $0.2$ & $0.9$ & $0.2$ & $0.9$ & $1.0$ & $0.0$ & $72.0$ & $18.7$ & $41.8$ & $5.6$ & $62.3$ & $16.9$ & $46.5$ & $13.8$ & $30.0$ & $6.0$ & $8.9$ \\

Darknet53~\cite{behley2019semantickitti} & $ 49.9 $ & $ 86.4 $ & $24.5 $ & $32.7$ & $ 25.5$ & $ 22.6 $ & $36.2 $ & $33.6 $ & $4.7$ & $ 91.8 $ & $64.8 $ & $74.6$ & $ 27.9 $ & $84.1$ & $ 55.0$ & $ 78.3 $ & $50.1$ & $ 64.0 $ & $38.9 $ & $ 52.2$ \\

RandLA-Net~\cite{hu2020randla} & $ 50.3  $ & $ 94.0 $ & $ 19.8 $ & $ 21.4 $ & $ 42.7 $ & $ 38.7 $ & $ 47.5 $ & $ 48.8 $ & $ 4.6  $ & $ 90.4 $ & $ 56.9 $ & $ 67.9 $ & $ 15.5 $ & $ 81.1 $ & $ 49.7 $ & $ 78.3 $ & $ 60.3 $ & $ 59.0 $ & $ 44.2 $ & $ 38.1 $ \\
RangeNet++~\cite{milioto2019rangenet++} & $ 52.2  $ & $ 91.4 $ & $ 25.7 $ & $ 34.4 $ & $ 25.7 $ & $ 23.0 $ & $ 38.3 $ & $  38.8 $ & $ 4.8 $ & $ 91.8 $ & $ 65.0 $ & $ 75.2 $ & $ 27.8 $ & $ 87.4 $ & $ 58.6 $ & $ 80.5 $ & $ 55.1 $ & $ 64.6 $ & $ 47.9 $ & $ 55.9$ \\
PolarNet~\cite{zhang2020polarnet} & $ 54.3  $ & $ 93.8 $ & $ 40.3 $ & $ 30.1 $ & $ 22.9 $ & $ 28.5 $ & $ 43.2 $ & $ 40.2 $ & $ 5.6 $ & $ 90.8 $ & $ 61.7 $ & $ 74.4 $ & $ 21.7 $ & $ 90.0 $ & $ 61.3 $ & $ 84.0 $ & $ 65.5 $ & $ 67.8 $ & $ 51.8 $ & $ 57.5 $ \\
SqueezeSegv3~\cite{xu2020squeezesegv3} & $ 55.9  $ & $ 92.5 $ & $ 38.7 $ & $ 36.5 $ & $ 29.6 $ & $ 33.0 $ & $ 45.6 $ & $ 46.2 $ & $ {20.1} $ & $ 91.7 $ & $ 63.4 $ & $ 74.8 $ & $ 26.4 $ & $ 89.0 $ & $ 59.4 $ & $ 82.0 $ & $ 58.7 $ & $ 65.4 $ & $ 49.6 $ & $ 58.9  $ \\

KPConv~\cite{thomas2019kpconv} & $ 58.8  $ & $ 96.0 $ & $ 32.0 $ & $ 42.5 $ & $ 33.4$ & $44.3$ & $61.5 $ & $ 61.6 $ & $ 11.8 $ & $ 88.8 $ & $ 61.3$ & $  72.7$ & $31.6$ & $ \mathbf{95.0} $ & $ 64.2 $ & $ 84.8 $ & $ 69.2 $ & $ 69.1 $ & $ 56.4 $ & $ 47.4 $ \\

Salsanext~\cite{cortinhal2020salsanext} & $ 59.5  $ & $ 91.9 $ & $ 48.3 $ & $ 38.6 $ & $ 38.9 $ & $ 31.9 $ & $ 60.2 $ & $ 59.0 $ & $ 19.4 $ & $ 91.7 $ & $ 63.7 $ & $ 75.8 $ & $ 29.1 $ & $ 90.2 $ & $ 64.2 $ & $ 81.8 $ & $ 63.6 $ & $ 66.5 $ & $ 54.3 $ & $ 62.1$ \\

FusionNet~\cite{zhang2020deep} & $ 61.3  $ & $ 95.3 $ & $ 47.5 $ & $ 37.7 $ & $ 41.8 $ & $ 34.5 $ & $ 59.5 $ & $ 56.8 $ & $ 11.9 $ & $ 91.8 $ & $ 68.8 $ & $ 77.1 $ & $ 30.8 $ & $ 92.5 $ & $ 69.4 $ & $ 84.5 $ & $ 69.8 $ & $ 68.5$ & $60.4 $ & $ 66.5$ \\ 
KPRNet~\cite{kochanov2020kprnet} & $ 63.1  $ & $ 95.5$ & $54.1$ & $ 47.9$ & $23.6 $ & $ 42.6$ & $65.9 $ & $ 65.0 $ & $ 16.5 $ & $ 93.2 $ & $ 73.9 $ & $ 80.6 $ & $ 30.2 $ & $ 91.7 $ & $ {68.4} $ & $ 85.7 $ & $ 69.8 $ & $ 71.2 $ & $ 58.7 $ & $ 64.1$ \\
TORNADONet~\cite{gerdzhev2021tornado} & $ 63.1 $ & $94.2$ & $ 55.7$ & $ 48.1$ & $ 40.0$ & $ 38.2$ & $ 63.6$ & $ 60.1$ & $ 34.9$ & $ 89.7$ & $ 66.3$ & $ 74.5$ & $ 28.7$ & $ 91.3$ & $ 65.6$ & $ 85.6$ & $ 67.0$ & $ 71.5 $ & $ 58.0 $ & $ {65.9}$ \\

RangeViT~\cite{ando2023rangevit} & $64.0$ & $95.4$ & $55.8$ & $43.5$ & $29.8$ & $42.1$ & $63.9$ & $58.2$ & $38.1$ & $93.1$ & $70.2$ & $80.0$ & $32.5$ & $92.0$ & $69.0 $ & $85.3$ & $70.6$ & $71.2$ & $60.8$ & $64.7$  \\

AMVNet~\cite{liong2020amvnet} & $ 65.3  $ & $ 96.2 $ & $ 59.9 $ & $ 54.2 $ & $ 48.8 $ & $ 45.7 $ & $ 71.0 $ & $ 65.7 $ & $ 11.0 $ & $ 90.1 $ & $ 71.0 $ & $ 75.8 $ & $ 32.4 $ & $ 92.4 $ & $ 69.1 $ & $ 85.6 $ & $ 71.7 $ & $  69.6 $ & $ 62.7 $ & $ 67.2 $ \\
GFNet~\cite{qiu2022GFNet} & $65.4$ & $96.0$ & $53.2$ & $48.3$ & $31.7$ & $47.3$ & $62.8$ & $57.3$ & $44.7$ & $\mathbf{93.6}$ & $72.5$ & $\mathbf{80.8}$ & $31.2$ & $94.0$ & $\mathbf{73.9}$ & $85.2$ & $71.1$ & $69.3$ & $61.8$ & $68.0$\\

JS3C-Net~\cite{yan2021sparse} & $ 66.0  $ & $ 95.8$ & $ 59.3$ & $ 52.9$ & $ 54.3 $ & $ 46.0 $ & $ 69.5 $ & $ 65.4 $ & $ 39.9 $ & $ 88.9 $ & $ 61.9 $ & $ 72.1 $ & $ 31.9 $ & $ 92.5 $ & $ 70.8 $ & $ 84.5 $ & $ 69.8 $ & $ 67.9 $ & $ 60.7 $ & $ 68.7 $  \\
SPVNAS~\cite{tang2020searching} & $ 66.4  $ & $ 97.3 $ & $ 51.5 $ & $ 50.8 $ & $ 59.8 $ & $ 58.8 $ & $ 65.7 $ & $ 65.2 $ & $ 43.7 $ & $ 90.2 $ & $ 67.6 $ & $ 75.2 $ & $ 16.9 $ & $ 91.3 $ & $ 65.9 $ & $ 86.1 $ & $ 73.4 $ & $ 71.0 $ & $ 64.2 $ & $ 66.9$ \\

WaffleIron~\cite{puy23waffleiron} & $67.3$ & $96.5$ & $62.3$ & $64.1$ & $55.2$ & $48.7$ & $70.4$ & $77.8$ & $29.6$ & $90.5$ & $69.5$ & $75.9$ & $24.6$ & $91.8$ & $68.1$ & $85.4$ & $70.8$ & $69.6$ & $62.0$ & $65.2$
\\
Cylinder3D~\cite{zhu2021cylindrical} & $ 68.9  $ & $ 97.1 $ & $ 67.6 $ & $ 63.8 $ & $ 50.8 $ & $ 58.5 $ & $ 73.7 $ & $ 69.2 $ & $ 48.0 $ & $ 92.2 $ & $ 65.0 $ & $ 77.0 $ & $ 32.3 $ & $ 90.7 $ & $ 66.5 $ & $ 85.6 $ & $ 72.5 $ & $ 69.8 $ & $ 62.4 $ & $ 66.2$ \\

AF2S3Net~\cite{af2s3net} & $ 69.7  $ & $ 94.5 $ & $ 65.4 $ & $ \mathbf{86.8} $ & $ 39.2 $ & $ 41.1 $ & $ \mathbf{80.7} $ & $ 80.4 $ & $ \mathbf{74.3} $ & $ 91.3 $ & $ 68.8 $ & $ 72.5 $ & $ \mathbf{53.5} $ & $ 87.9 $ & $ 63.2 $ & $ 70.2 $ & $ 68.5 $ & $ 53.7 $ & $ 61.5 $ & $ 71.0$ \\

RPVNet~\cite{rpvnet} & $ 70.3  $ & $ 97.6 $ & $ 68.4 $ & $ 68.7 $ & $ 44.2 $ & $ 61.1 $ & $ 75.9 $ & $ 74.4 $ & $ 73.4 $ & $ 93.4 $ & $ 70.3 $ & $ 80.7 $ & $ 33.3 $ & $ 93.5 $ & $ 72.1 $ & $ 86.5 $ & $ 75.1 $ & $ 71.7 $ & $ 64.8 $ & $ 61.4$ \\

SDSeg3D~\cite{li2022self} & $ 70.4  $ & $ 97.4 $ & $ 58.7 $ & $ 54.2 $ & $ 54.9 $ & $ 65.2 $ & $ 70.2 $ & $ 74.4 $ & $ 52.2 $ & $ 90.9 $ & $ 69.4 $ & $ 76.7 $ & $  41.9 $ & $ 93.2 $ & $ 71.1 $ & $ 86.1 $ & $ 74.3 $ & $ 71.1 $ & $ 65.4 $ & $ 70.6 $ \\

GASN~\cite{ye2022efficient} & $ 70.7  $ & $ 96.9 $ & $ 65.8 $ & $  58.0 $ & $ 59.3 $ & $  61.0 $ & $ 80.4 $ & $ \mathbf{82.7} $ & $ 46.3 $ & $ 89.8 $ & $ 66.2 $ & $ 74.6 $ & $ 30.1 $ & $ 92.3 $ & $ 69.6 $ & $ 87.3 $ & $ 73.0 $ & $ 72.5 $ & $ 66.1 $ & $ \mathbf{71.6}$ \\
PVKD~\cite{pvkd2022} & $ 71.2 $ & $ 97.0 $ & $ 67.9 $ & $ 69.3 $ & $ 53.5 $ & $ 60.2 $ & $ 75.1 $ & $ 73.5 $ & $ 50.5 $ & $ 91.8 $ & $ 70.9 $ & $ 77.5 $ & $ 41.0 $ & $ 92.4 $ & $ 69.4 $ & $ 86.5 $ & $ 73.8 $ & $ 71.9 $ & $ 64.9 $ & $ 65.8$ \\
2DPASS~\cite{yan20222dpass} & $ 72.9  $ & $ 97.0 $ & $ 63.6 $ & $ 63.4 $ & $ 61.1 $ & $ 61.5 $ & $ 77.9 $ & $ 81.3 $ & $ 74.1 $ & $ 89.7 $ & $ 67.4 $ & $ 74.7 $ & $ 40.0 $ & $ 93.5 $ & $ 72.9 $ & $ 86.2 $ & $ 73.9 $ & $ 71.0 $ & $ 65.0 $ & $ 70.4$ \\
RangeFormer~\cite{kong2023rethinking} & $73.3$ & $96.7$ & $69.4$ & $73.7$ & $59.9$ & $66.2$ & $78.1$ & $75.9$ & $58.1$ & $92.4$ & $73.0$ & $78.8$ & $42.4$ & $92.3$ & $70.1$ & $86.6$ & $73.3$ & $72.8$ & $66.4$ & $66.6$
\\
\midrule
\cellcolor{cyan!9}\textbf{\algorithmname~(Ours)} & \cellcolor{cyan!9}$\mathbf{75.2}$ & \cellcolor{cyan!9}$\mathbf{97.9}$ & \cellcolor{cyan!9}$\mathbf{71.9}$ & \cellcolor{cyan!9}$75.2$ & \cellcolor{cyan!9}$\mathbf{63.6}$ & \cellcolor{cyan!9}$\mathbf{74.1}$ & \cellcolor{cyan!9}$78.9$ & \cellcolor{cyan!9}$74.8$ & \cellcolor{cyan!9}$60.6$ & \cellcolor{cyan!9}$92.6$ & \cellcolor{cyan!9}$\mathbf{74.0}$ & \cellcolor{cyan!9}$79.5$ & \cellcolor{cyan!9}$46.1$ & \cellcolor{cyan!9}$93.4$ & \cellcolor{cyan!9}$72.7$ & \cellcolor{cyan!9}$\mathbf{87.5}$ & \cellcolor{cyan!9}$\mathbf{76.3}$ & \cellcolor{cyan!9}$\mathbf{73.1}$ & \cellcolor{cyan!9}$\mathbf{68.3}$ & \cellcolor{cyan!9}$68.5$
\\
\bottomrule
\end{tabular}
\end{adjustbox}
\vspace{0.2cm}
\end{table*}
\begin{table*}[ht]
\caption{Quantitative results of UniSeg and state-of-the-art \textbf{LiDAR panoptic segmentation} methods on the \textit{test} set of \textbf{SemanticKITTI}~\cite{behley2019semantickitti}.}
\vskip -0.2cm
\centering\scalebox{0.77}{
\begin{tabular}{r|cccc|ccc|ccc|c}
\toprule
Methods & \PQ & \PQda & \RQ & \SQ & \PQth & \RQth & \SQth & \PQst & \RQst & \SQst & \miou
\\\midrule\midrule
RangeNet++ \cite{milioto2019rangenet++} +
PointPillars \cite{lang2019pointpillars}  & $ 37.1          $ & $ 45.9          $ & $ 47.0          $ & $ 75.9          $ & $ 20.2          $ & $ 25.2          $ & $ 75.2          $ & $ 49.3          $ & $ 62.8          $ & $ 76.5                  $ & $ 52.4$
\\
LPASD \cite{milioto2020lidar}  & $ 38.0          $ & $ 47.0          $ & $ 48.2          $ & $ 76.5          $ & $ 25.6          $ & $ 31.8          $ & $ 76.8          $ & $ 47.1          $ & $ 60.1          $ & $ 76.2                  $ & $ 50.9$
\\
KPConv \cite{thomas2019kpconv} + PointPillars \cite{lang2019pointpillars}   & $ 44.5          $ & $ 52.5          $ & $ 54.4          $ & $ 80.0          $ & $ 32.7          $ & $ 38.7          $ & $ 81.5          $ & $ 53.1          $ & $ 65.9          $ & $ 79.0                  $ & $ 58.8$
\\
SalsaNext \cite{cortinhal2020salsanext} +
PV-RCNN \cite{shi2020pv}  & $ 47.6 $ & $ 55.3 $ & $ 58.6 $ & $ 79.5 $ & $ 39.1 $ & $ 45.9 $ & $ 82.3 $ & $ 53.7 $ & $ 67.9 $ & $ 77.5 $ & $ 58.9$
\\
KPConv~\cite{thomas2019kpconv} + PV-RCNN~\cite{shi2020pv}  & $ 50.2  $ & $ 57.5 $ & $ 61.4 $ & $ 80.0 $ & $ 43.2 $ & $ 51.4 $ & $ 80.2 $ & $ 55.9 $ & $ 68.7 $ & $ 79.9 $ & $ 62.8$
\\
Panoster \cite{gasperini2021panoster}        & $ 52.7 $ & $ 59.9 $ & $ 64.1 $ & $ 80.7 $ & $ 49.9 $ & $ 58.8 $ & $ 83.3 $ & $ 55.1 $ & $ 68.2 $ & $ 78.8 $ & $ 59.9$ \\
Panoptic-PolarNet~\cite{zhou2021panoptic}& $54.1 $ & $ 60.7 $ & $ 65.0 $ & $ 81.4 $ & $ 53.3 $ & $ 60.6 $ & $ 87.2 $ & $ 54.8 $ & $ 68.1 $ & $ 77.2 $ & $ 59.5  $ 
\\
DS-Net~\cite{dsnet}  & $ 55.9 $ & $ 62.5 $ & $ 66.7 $ & $ 82.3 $ & $ 55.1 $ & $ 62.8 $ & $ 87.2 $ & $ 56.5 $ & $ 69.5 $ & $ 78.7 $ & $ 61.6$
\\
EfficientLPS~\cite{sirohi2021efficientlps}  & $ 57.4 $ & $ 63.2 $ & $  68.7 $ & $ 83.0  $ & $ 53.1 $ & $ 60.5 $ & $ 87.8$ & $ 60.5 $ & $ 74.6 $ & $ 79.5 $ & $ 61.4$  \\
GP-S3Net~\cite{razani2021gp}  & $ 60.0 $ & $ 69.0 $ & $ 72.1 $ & $ 82.0 $ & $ 65.0 $ & $ 74.5 $ & $ 86.6 $ & $ 56.4 $ & $ 70.4 $ & $ 78.7 $ & $ 70.8$
\\
SCAN~\cite{xu2022sparse}  & $ 61.5 $ & $ 67.5 $ & $ 72.1 $ & $ 84.5 $ & $ 61.4 $ & $  69.3 $ & $ 88.1 $ & $ 61.5 $ & $ 74.1 $ & $ 81.8 $ & $ 67.7$
\\
Panoptic-PHNet~\cite{li2022panoptic}  & $ 64.6 $ & $ 70.2 $ & $ 74.9 $ & $ \mathbf{85.7} $ & $ 66.9 $ & $ 73.3 $ & $ \mathbf{91.5} $ & $ 63.0 $ & $ 76.1 $ & $ 81.5 $ & $ 68.4$
\\\midrule
\cellcolor{cyan!9}\textbf{\algorithmname~(Ours)} & \cellcolor{cyan!9}$\mathbf{67.2}$ & \cellcolor{cyan!9}$\mathbf{72.1}$ & \cellcolor{cyan!9}$\mathbf{78.1} $& \cellcolor{cyan!9}$85.5$ & \cellcolor{cyan!9}$\mathbf{67.5}$ & \cellcolor{cyan!9}$\mathbf{75.7}$ & \cellcolor{cyan!9}$89.0$ & \cellcolor{cyan!9}$\mathbf{67.0}$ & \cellcolor{cyan!9}$\mathbf{79.8} $& \cellcolor{cyan!9}$\mathbf{83.0}$ & \cellcolor{cyan!9}$\mathbf{73.8}$
\\\bottomrule
\end{tabular}
}
\label{tab:sem_ps_sup}
\vspace{0.2cm}
\end{table*}

\begin{table*}[!t]
\caption{Quantitative results of \algorithmname~and state-of-the-art \textbf{LiDAR semantic segmentation} methods on the \textit{test} set of \textbf{nuScenes}~\cite{caesar2020nuscenes}.}
\vskip -0.2cm
\label{tab:nusc_sup}
\centering
\begin{adjustbox}{width=\textwidth}
\begin{tabular}{r|c|c|c|c|c|c|c|c|c|c|c|c|c|c|c|c|c}
\toprule
\textbf{Model} & \rotatebox{90}{\textbf{mIoU}} & \rotatebox{90}{barrier} &  \rotatebox{90}{bicycle} & \rotatebox{90}{bus} & \rotatebox{90}{car} & \rotatebox{90}{construction} & \rotatebox{90}{motorcycle} & \rotatebox{90}{pedestrian} & \rotatebox{90}{traffic-cone} & \rotatebox{90}{trailer} & \rotatebox{90}{truck} & \rotatebox{90}{driveable} & \rotatebox{90}{other} &
\rotatebox{90}{sidewalk} & \rotatebox{90}{terrain} & \rotatebox{90}{manmade} & \rotatebox{90}{vegetation} \\
\midrule
\midrule

PolarNet~\cite{zhang2020polarnet}  & $ 69.4 $ & $  72.2 $ & $ 16.8 $ & $ 77.0 $ & $ 86.5 $ & $ 51.1 $ & $ 69.7 $ & $  64.8 $ & $ 54.1 $ & $ 69.7 $ & $ 63.5 $ & $ 96.6 $ & $ 67.1 $ & $ 77.7 $ & $ 72.1 $ & $ 87.1 $ & $ 84.5$ \\

JS3C-Net~\cite{yan2021sparse}  & $ 73.6  $ & $ 80.1  $ & $ 26.2 $ & $ 87.8 $ & $ 84.5 $ & $ 55.2 $ & $ 72.6 $ & $ 71.3 $ & $ 66.3 $ & $ 76.8 $ & $ 71.2 $ & $ 96.8 $ & $ 64.5 $ & $ 76.9 $ & $ 74.1 $ & $ 87.5 $ & $ 86.1$ \\

PMF~\cite{zhuang2021perception}
 & $77.0 $ & $ 82.0$ & $ 40.0$ & $ 81.0$ & $ 88.0$ & $ 64.0 $ & $79.0$ & $ 80.0 $ & $ 76.0$ & $ 81.0$ & $ 67.0$ & $ 97.0$ & $ 68.0$ & $ 78.0$ & $ 74.0$ & $ 90.0$ & $ 88.0$\\

Cylinder3D~\cite{zhu2021cylindrical}  & $ 77.2 $ & $ 82.8 $ & $ 29.8 $ & $ 84.3 $ & $ 89.4 $ & $ 63.0 $ & $ 79.3 $ & $ 77.2 $ & $ 73.4 $ & $ 84.6 $ & $ 69.1 $ & $ 97.7 $ & $ 70.2 $ & $ 80.3 $ & $ 75.5 $ & $ 90.4 $ & $ 87.6$  \\

AMVNet~\cite{liong2020amvnet}  & $77.3 $ & $  80.6 $ & $ 32.0 $ & $ 81.7 $ & $ 88.9 $ & $ 67.1 $ & $ 84.3 $ & $ 76.1 $ & $ 73.5 $ & $ 84.9 $ & $ 67.3 $ & $ 97.5 $ & $ 67.4 $ & $ 79.4 $ & $ 75.5 $ & $ 91.5 $ & $ 88.7$\\

SPVCNN~\cite{tang2020searching}  & $77.4$ & $ 80.0$ & $ 30.0 $ & $ 91.9 $ & $ 90.8 $ & $ 64.7 $ & $ 79.0 $ & $ 75.6 $ & $ 70.9 $ & $ 81.0 $ & $ 74.6 $ & $ 97.4 $ & $ 69.2 $ & $ 80.0 $ & $ 76.1 $ & $ 89.3 $ & $ 87.1$ \\

AF2S3Net~\cite{af2s3net}  & $78.3 $ & $ 78.9 $ & $ 52.2 $ & $ 89.9 $ & $ 84.2 $ & $ 77.4 $ & $ 74.3 $ & $ 77.3 $ & $ 72.0 $ & $ 83.9 $ & $ 73.8 $ & $ 97.1 $ & $ 66.5 $ & $ 77.5 $ & $ 74.0 $ & $ 87.7 $ & $ 86.8$ \\

2D3DNet~\cite{genova2021learning}  & $ 80.0 $ & $ 83.0 $ & $ 59.4 $ & $ 88.0 $ & $85.1 $ & $ 63.7 $ & $ 84.4 $ & $ 82.0 $ & $ 76.0 $ & $ 84.8 $ & $ 71.9 $ & $ 96.9 $ & $ 67.4 $ & $ 79.8 $ & $ 76.0 $ & $ \mathbf{92.1} $ & $ 89.2 $\\

GASN~\cite{ye2022efficient}  & $80.4 $ & $ 85.5 $ & $ 43.2 $ & $ 90.5 $ & $ \mathbf{92.1} $ & $ 64.7 $ & $ 86.0 $ & $ 83.0 $ & $ 73.3 $ & $ 83.9 $ & $ 75.8 $ & $ 97.0 $ & $ 71.0 $ & $ \mathbf{81.0} $ & $ \mathbf{77.7} $ & $ 91.6 $ & $ \mathbf{90.2}$\\

2DPASS~\cite{yan20222dpass}  & $ 80.8 $ & $ 81.7 $ & $ 55.3 $ & $ 92.0 $ & $ 91.8 $ & $ 73.3 $ & $ 86.5 $ & $ 78.5 $ & $ 72.5 $ & $ 84.7 $ & $ 75.5 $ & $ 97.6 $ & $ 69.1 $ & $ 79.9 $ & $ 75.5 $ & $ 90.2 $ & $ 88.0$\\

LidarMultiNet~\cite{lidarmultinet}  & $ 81.4 $ & $ 80.4 $ & $ 48.4 $ & $ \mathbf{94.3} $ & $ 90.0 $ & $ 71.5 $ & $ 87.2 $ & $ \mathbf{85.2} $ & $ \mathbf{80.4} $ & $ \mathbf{86.9} $ & $ 74.8 $ & $ \mathbf{97.8} $ & $ 67.3 $ & $ 80.7 $ & $ 76.5 $ & $ \mathbf{92.1} $ & $ 89.6$ \\
\midrule
\cellcolor{cyan!9}\textbf{\algorithmname~(Ours)} &\cellcolor{cyan!9} $ \mathbf{83.5} $ &\cellcolor{cyan!9} $ \mathbf{85.9} $ &\cellcolor{cyan!9} $ \mathbf{71.2} $ &\cellcolor{cyan!9} $ 92.1 $ &\cellcolor{cyan!9} $ 91.6 $ &\cellcolor{cyan!9} $ \mathbf{80.5} $ &\cellcolor{cyan!9} $ \mathbf{88.0} $ &\cellcolor{cyan!9} $ 80.9 $ &\cellcolor{cyan!9} $ 76.0 $ &\cellcolor{cyan!9} $ 86.3 $ &\cellcolor{cyan!9} $ \mathbf{76.7} $ &\cellcolor{cyan!9} $ 97.7 $ &\cellcolor{cyan!9} $ \mathbf{71.8} $ &\cellcolor{cyan!9} $ 80.7 $ &\cellcolor{cyan!9} $ 76.7 $ &\cellcolor{cyan!9} $  91.3 $ &\cellcolor{cyan!9} $ 88.8$ \\
\bottomrule
\end{tabular}
\end{adjustbox}
\vspace{0.2cm}
\end{table*}

\begin{table*}[t]
\caption{Quantitative results of UniSeg and state-of-the-art \textbf{LiDAR semantic segmentation} methods on the \textit{val} set of \textbf{Waymo Open Dataset}~\cite{waymo}. Methods with * are our implementations.}
\vskip -0.2cm
\label{tab:waymo_sup}
\centering
\begin{adjustbox}{width=\textwidth}
\begin{tabular}{r|c|c|c|c|c|c|c|c|c|c|c|c|c|c|c|c|c|c|c|c|c|c|c}
\toprule
\textbf{Model} & \rotatebox{90}{\textbf{mIoU}} & \rotatebox{90}{car} &  
\rotatebox{90}{truck} &  
\rotatebox{90}{bus} & 
\rotatebox{90}{other vehicle} &  \rotatebox{90}{motorcyclist} &  \rotatebox{90}{bicyclist} &  \rotatebox{90}{pedestrian} & 
\rotatebox{90}{sign} & 
\rotatebox{90}{traffic light} &  
\rotatebox{90}{pole} & 
\rotatebox{90}{construction} &  \rotatebox{90}{bicycle} &
\rotatebox{90}{motorcycle} &  \rotatebox{90}{building} &  \rotatebox{90}{vegetation} & 
\rotatebox{90}{tree trunk} &  
\rotatebox{90}{curb} & 
\rotatebox{90}{road} & 
\rotatebox{90}{lane marker} & 
\rotatebox{90}{other ground} & \rotatebox{90}{walkable} &
\rotatebox{90}{sidewalk}
\\
\midrule
\midrule

P-Transformer*~\cite{point-transformer}  & $ {63.3} $ & $ 93.1 $ & $ 58.8 $ & $ 61.4 $ & $ 25.4 $ & $ 0.0 $ & $ 67.9 $ & $ 85.5 $ & $ 72.3 $ & $ 36.2 $ & $ 71.4 $ & $ 66.4 $ & $ 58.7 $ & $ 54.3 $ & $ 93.7 $ & $ 90.0 $ & $ 64.7 $ & $ 65.2 $ & $ 90.4 $ & $ 48.2 $ & $ 42.8 $ & $ 74.5 $ & $ 71.7$  \\
Cylinder3D*~\cite{zhu2021cylindrical}  & $ {66.0} $ & $ \mathbf{95.1} $ & $ 59.6 $ & $ 74.1 $ & $ 28.7 $ & $ \mathbf{2.4} $ & $ 62.3 $ & $ 86.8 $ & $ 71.5 $ & $ 33.6 $ & $ 73.4 $ & $ 65.2 $ & $ 62.0 $ & $ 76.5 $ & $ 95.1 $ & $ \mathbf{91.0} $ & $ 66.6 $ & $ 65.5 $ & $ 92.3 $ & $ 49.9 $ & $ 47.1 $ & $ \mathbf{79.0} $ & $ 75.1$  \\

SPVCNN*~\cite{tang2020searching} & {67.4} & 94.3 & 59.8 & 78.5 & 27.5 & 0.0 & 70.8 & 87.8 & 74.9 & 39.2 & 74.4 & 69.5 & 70.4 & 79.4 & 94.8 & 90.8 & 66.9 & 66.6 & 91.7 & 50.9 & 43.9 & 77.2 & 72.7 \\

\midrule
\cellcolor{cyan!9}\textbf{\algorithmname~(Ours)} & \cellcolor{cyan!9}$\mathbf{{69.6}}$ & \cellcolor{cyan!9}$94.4$ & \cellcolor{cyan!9}$\mathbf{60.4}$ & \cellcolor{cyan!9}$\mathbf{79.6}$ & \cellcolor{cyan!9}$\mathbf{40.6} $& \cellcolor{cyan!9}$0.0$ & \cellcolor{cyan!9}$\mathbf{73.2}$ & \cellcolor{cyan!9}$\mathbf{89.0} $& \cellcolor{cyan!9}$\mathbf{75.7}$ & \cellcolor{cyan!9}$\mathbf{43.3} $& \cellcolor{cyan!9}$\mathbf{76.1} $& \cellcolor{cyan!9}$\mathbf{70.2} $& \cellcolor{cyan!9}$\mathbf{75.5}$ & \cellcolor{cyan!9}$\mathbf{80.8}$ & \cellcolor{cyan!9}$\mathbf{95.2} $& \cellcolor{cyan!9}$\mathbf{91.0}$ & \cellcolor{cyan!9}$\mathbf{68.2} $& \cellcolor{cyan!9}$\mathbf{68.7}$ & \cellcolor{cyan!9}$\mathbf{92.6} $& \cellcolor{cyan!9}$\mathbf{53.9} $& \cellcolor{cyan!9}$\mathbf{48.3}$ & \cellcolor{cyan!9}$78.8$ & \cellcolor{cyan!9}$\mathbf{75.8}$ \\
\bottomrule
\end{tabular}
\end{adjustbox}
\vspace{0.2cm}
\end{table*}

\section{Additional Implementation Details}
\label{sec:implementation}
\noindent \textbf{Network Structure}. For the image branch, the input image size is 376$\times$1241 on the SemanticKITTI~\cite{behley2019semantickitti} dataset. For the multi-camera images of nuScenes~\cite{caesar2020nuscenes,panoptic-nuscenes} and Waymo Open~\cite{waymo} datasets, the image size is 900$\times$1600 and 640$\times$960, respectively. For the range branch,  the input range-image size on the SemanticKITTI, nuScenes and Waymo Open datasets are 64$\times$2048, 32$\times$1920, and 64$\times$2688, respectively.
To construct a robust point-voxel-range fusion network for the point cloud branch, we first construct the point-voxel backbone based on the Minkowski-UNet34~\cite{choy20194d}. Then, we add the range-image branch, i.e., SalsaNext~\cite{cortinhal2020salsanext}, to the point-voxel network and perform point-voxel-range fusion by the \textbf{L}earnable cross-\textbf{V}iew \textbf{A}ssociation module (LVA). Range and voxel branches are UNet-like architectures with four down-sampling stages and four up-sampling stages. The dimensions of these nine stages are 32, 32, 64, 128, 256, 256, 128, 96, and 96, respectively, and the point branch includes 4 MLPs with channel dimensions being 32, 256, 128, and 96, respectively. In addition, to increase model capacity, the channel expansion ratio is set as 1.75, 1.6, 1.6 for SemanticKITTI, nuScenes and Waymo Open datasets, respectively. 
We use ImageNet-pretrained ResNet-34~\cite{he2016deep} as the feature extractor for the image backbone. The image backbone can be flexibly selected from off-the-shelf networks.

\noindent \textbf{Data Augmentation and Test-Time Augmentation}. We take different data augmentation strategies for the point cloud and image branches. For the image branch, we do not perform data augmentation. For the point cloud branch, we perform random flip $(\tau_{flip})$ along with the $X$ axis, $Y$ axis and $XY$ axis, and random translation $(\tau_{trans})$ within the normal distribution of $[0, 0.1]$ as well as LaserMix~\cite{lasermix} and PolarMix~\cite{xiao2022polarmix}. Global scaling $(\tau_{scal})$ and global rotation $(\tau_{rot})$ are also adopted. The scaling factor and rotation angle are randomly selected within $[0.9, 1.1]$ and $[0, 2\pi]$ for random scaling and random rotation. 
To further improve the performance of our model on the online leaderboard,  we fine-tune our trained model on both train and validation set for 12 or 24 epochs with cosine annealing schedule~\cite{loshchilov2016sgdr} on the SemanticKITTI and nuScenes datasets, respectively, and adopt new Test-Time Augmentation (TTA) strategy as in ~\cite{li2022self}. Specifically, given an input LiDAR scan $\mathbf{p}\in\mathbb{R}^{N\times3}$ in a LiDAR point cloud with coordinates $(p^x, p^y, p^z)$. We apply the above four data augmentation transformations for $\mathbf{p}$ in a compound way  $\tau_{comp}(\mathbf{p}) = \tau_{trans}(\tau_{flip}(\tau_{scal}(\tau_{rot}(\mathbf{p}))) )$. The input scan is augmented into a set of $\{\mathbf{p}, \mathbf{p}_{comp, i} \}$, where $i$ is the index of the augmented samples in the set. After that, the output of the prediction from multiple augmented of input LiDAR scan  $\mathbf{p}$ are summed and performed the \textit{argmax} to generate the final predictions at the inference stage. Note that the rotating angles are $\{0,\pm \frac{\pi}{8},   \pm \frac{\pi}{4}, \pm \frac{3\pi}{4},   \pm \frac{7\pi}{8}, \pi\}$ for yaw rotation in test-time.

\noindent \textbf{Panoptic Head}.  We follow the instance head design in~\cite{panopticpolarnet} to predict the instance centers and offsets for each BEV pixel. During the training phase,  we encode the ground-truth center map by a 2D Gaussian distribution around each instance's mass center and create an offset map where the offset measures the distance to its corresponding instance's mass center. The size of the center map and the offset map is 480$\times$360. The semantic segmentation predictions are utilized to create the foreground mask to form instance groups. Then, we conduct 2D class-agnostic instance grouping by predicting the center heatmap and offset for each point on the $XY$-plane. Finally, each instance group is assigned a unique label via majority voting to create the final panoptic segmentation. For the nuScenes panoptic segmentation, we follow~\cite{lidarmultinet} to refine the instance segmentation results via the predicted bounding boxes of the TransFusion detector~\cite{transfusion}. 
For the panoptic segmentation evaluation, we evaluate the predicted instance with a minimal point of 30, and 50 as a valid instance on the nuScenes and SemanticKITTI datasets, respectively.

\noindent \textbf{Evaluation Metrics}. The definition of Panoptic Quality (PQ)~\cite{kirillov2019panoptic}, Segmentation Quality (SQ), and Recognition Quality (RQ) is given as follows:
\begin{equation}
    \text{PQ} = \underbrace{\frac{\sum_{(i, j)\in TP}\text{IoU}(i, j)}{|TP|}}_\text{SQ} \times \underbrace{\frac{|TP|}{|TP| + \frac{1}{2}|FP| + \frac{1}{2}|FN|}}_\text{RQ}.
\end{equation}
The aforementioned three metrics are also calculated separately on \things{} and \stuff{} classes which produce
\PQth{}, \SQth{}, \RQth{}, and \PQst{}, \SQst{}, \RQst{}.
In addition, we report \PQda{} which is defined by swapping \PQ{} of each \stuff{} class to its IoU and then averaging over all classes. 
\section{Additional Quantitative Result}
\label{sec:quan}

We provide a more comprehensive comparison between \algorithmname~and competitive LiDAR segmentation networks.
Table~\ref{tab:sem_ss_sup} shows the
class-wise IoU scores of different LiDAR semantic segmentation methods on the \textit{test set} of SemanticKITTI~\cite{behley2019semantickitti}. Among all the LiDAR segmentation algorithms, \algorithmname~achieves compelling results. Table~\ref{tab:sem_ps_sup} shows the PQ, RQ, SQ, mIoU scores of different LiDAR panoptic segmentation methods on the \textit{test set} of SemanticKITTI~\cite{behley2019semantickitti}. We can observe a clear advantage of  \algorithmname~over other solutions. Table~\ref{tab:nusc_sup} shows the
class-wise IoU scores of different LiDAR semantic segmentation methods on the \textit{test set} of nuScenes~\cite{panoptic-nuscenes, caesar2020nuscenes}. \algorithmname~yields high mIoU scores than the SoTA solution of LidarMultiNet~\cite{lidarmultinet}, which demonstrates again the advantage of \algorithmname. In addition, we provide detailed performance on the Waymo Open~\cite{waymo} \textit{val set} in Table~\ref{tab:waymo_sup}. It shows \algorithmname~obtains higher efficacy.  
\section{Additional Qualitative Result}
\label{sec:qual}

\begin{figure*}[t]
 \centering
 \includegraphics[width=1.0\linewidth]{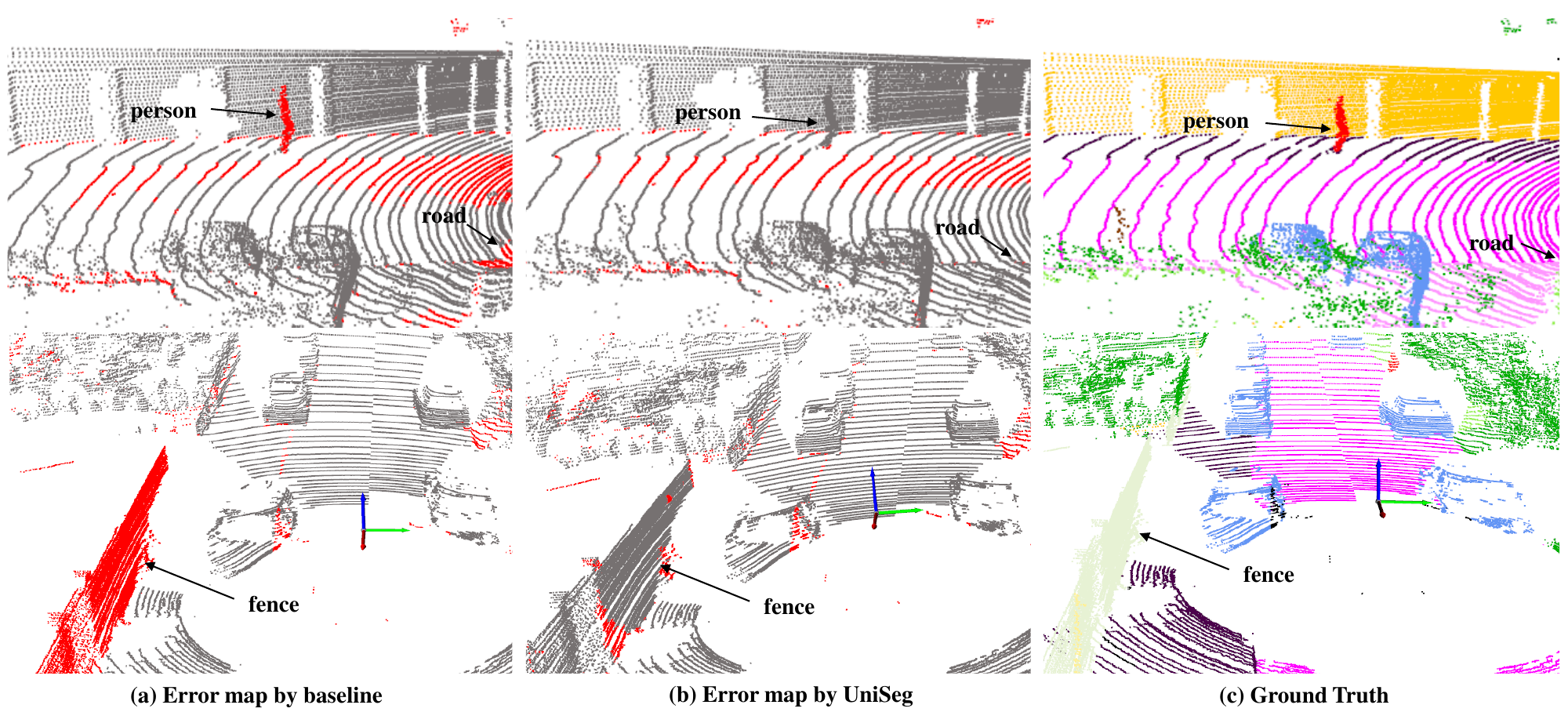}
 \vskip -0.3cm
 \caption{Qualitative results of UniSeg on the SemanticKITTI validation set.}
 \centering
 \label{fig:more_visual_compare}
\end{figure*}

\begin{figure*}[t]
 \centering
 \includegraphics[width=1.0\linewidth]{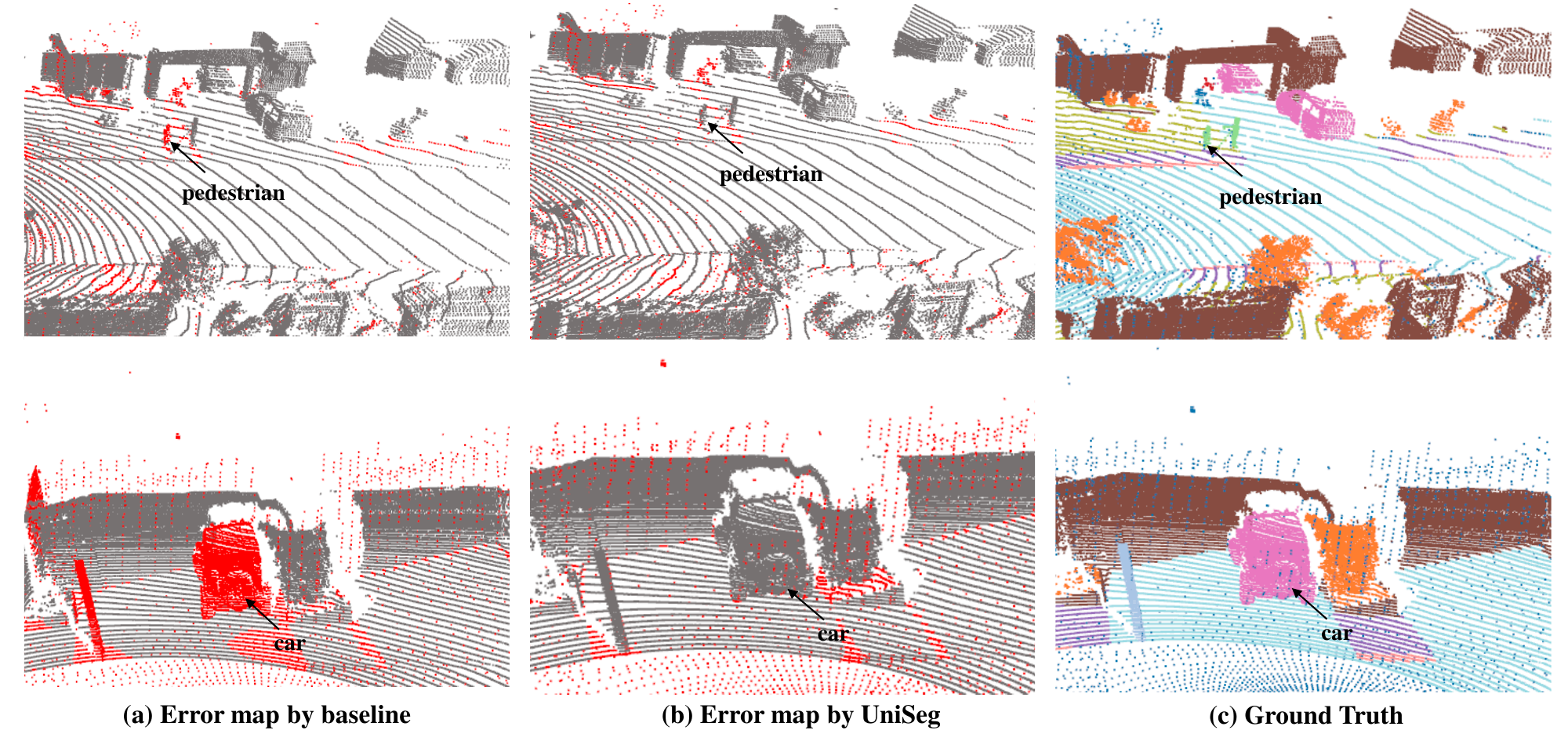}
 \vskip -0.3cm
 \caption{Qualitative results of UniSeg on the Waymo Open validation set.}
 \centering
 \label{fig:more_visual_comparison_waymo}
\end{figure*}

\begin{figure*}[t]
 \centering
 \includegraphics[width=1.0\linewidth]{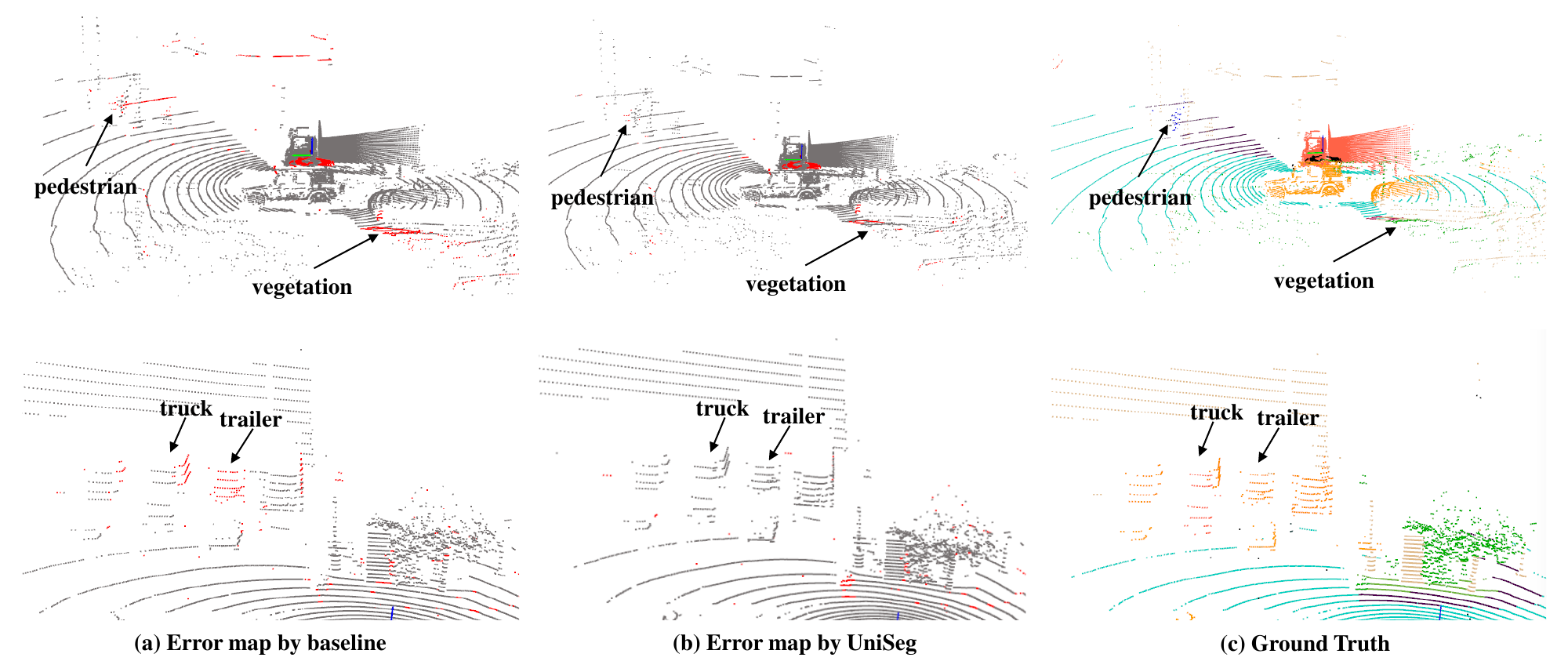}
 \vskip -0.3cm
 \caption{Qualitative results of UniSeg on the nuScenes validation set.}
 \centering
 \label{fig:more_visual_comparison_nuScenes}
\end{figure*}

We provide more visual comparisons of \algorithmname~with baseline algorithm (single modal) in Fig.~\ref{fig:more_visual_compare}, Fig.~\ref{fig:more_visual_comparison_waymo}, and Fig.~\ref{fig:more_visual_comparison_nuScenes} on the validation set of SemanticKITTI~\cite{behley2019semantickitti} , nuScenes~\cite{panoptic-nuscenes,caesar2020nuscenes}  and Waymo Open~\cite{waymo}, respectively. To highlight the differences in the error map, the correct/incorrect predictions are painted in gray/red, respectively. For the ground truth, different colors represent different classes. From Fig.~\ref{fig:more_visual_compare}, Fig.~\ref{fig:more_visual_comparison_waymo}, and Fig.~\ref{fig:more_visual_comparison_nuScenes}, the single-modal baseline has higher prediction errors than our \algorithmname, especially on small objects, \eg, pedestrians. For example, in Fig.~\ref{fig:more_visual_compare}, the baseline mistakenly predicts the person and fence and has higher prediction errors on the road boundaries. By contrast, \algorithmname~makes much better predictions on both person and fence, as well as the road boundaries, which is attributed to the comprehensive information provided by camera images and all views of the point cloud. In a nutshell, \algorithmname~can make more accurate point-wise predictions regardless of the distance and point density variation than the baseline.

\clearpage

{\small
\bibliographystyle{ieee_fullname}

\bibliography{egbib}
}

\end{document}